\documentclass[11pt]{article}
\usepackage[preprint]{acl}
\usepackage{multirow} 
\usepackage{times}
\usepackage{latexsym}
\usepackage{amsmath,amssymb}
\usepackage{graphicx}
\usepackage{hyperref}
\usepackage{enumitem}
\usepackage{algorithm}
\usepackage{algpseudocode}

\algrenewcommand\algorithmiccomment[1]{\hfill{\scriptsize\textit{#1}}}
\usepackage[T1]{fontenc}
\usepackage[utf8]{inputenc}
\usepackage{inconsolata}

\usepackage{tikz}
\usetikzlibrary{positioning}
\usepackage[table]{xcolor}
\definecolor{motifbg}{HTML}{E1F5EE}
\definecolor{motiffg}{HTML}{085041}
\definecolor{structbg}{HTML}{E6F1FB}
\definecolor{structfg}{HTML}{0C447C}
\definecolor{threshbg}{HTML}{FAEEDA}
\definecolor{threshfg}{HTML}{633806}
\definecolor{spanbg}{HTML}{E6F1FB}
\definecolor{spanfg}{HTML}{0C447C}
\definecolor{compbg}{HTML}{E1F5EE}
\definecolor{compfg}{HTML}{085041}
\newcommand{\pspan}[1]{\fcolorbox{spanfg}{spanbg}{\textcolor{spanfg}{\texttt{#1}}}}
\newcommand{\pcomp}[1]{\fcolorbox{compfg}{compbg}{\textcolor{compfg}{\texttt{#1}}}}
\newcommand{\pmotif}[1]{\colorbox{motifbg}{\textcolor{motiffg}{\texttt{#1}}}}

\definecolor{motifbg}{HTML}{FAEEDA}
\definecolor{motiffg}{HTML}{633806}
\usepackage{xcolor}
\definecolor{headcol}{HTML}{374151}
\definecolor{motifcol}{HTML}{6D28D9}
\definecolor{bodycol}{HTML}{1D4ED8}
\newcommand{\headpred}[1]{\textcolor{headcol}{\texttt{#1}}}
\newcommand{\bodypred}[1]{\textcolor{bodycol}{\texttt{#1}}}

\usepackage{booktabs}
\title{From Circuit Evidence to Mechanistic Theory:\\ An Inductive Logic Approach}
\author{Nura Aljaafari$^{1}$,~ Danilo S. Carvalho$^{3}$,~ Andr\'{e} Freitas$^{1,2,3}$ \\
  $^{1}$ Department of Computer Science, University of Manchester, United Kingdom\\
  $^{2}$ Idiap Research Institute, Switzerland\\
  $^{3}$ CRUK National Biomarker Centre, University of Manchester, United Kingdom\\
  \texttt{\{firstname.lastname\}@manchester.ac.uk}}
\date{}

\begin{document}
\maketitle

\begin{abstract}
Mechanistic interpretability produces circuit-level causal analyses of neural network behaviour, but discovered circuits often remain isolated experimental artefacts: there is no shared formal representation for what circuits compute, how they relate, or when two findings provide evidence for the same mechanism. This work provides a formal infrastructure for cumulative mechanistic science by treating circuit interpretation as \emph{inductive theory construction}. Each circuit is characterised at two levels: a \emph{Causal Functional Signature} (CFS), which grounds component behaviour in causal attribution evidence and token role profiles, and an \emph{architectural signature} $\tau_{\mathrm{arch}}$, learned by inductive logic programming (ILP) from scale-invariant structural predicates. Together, these constitute a formal coherence layer that makes mechanistic claims explicit, comparable via $\theta$-subsumption, and portable across model scales. CFS reveals qualitatively distinct computational strategies across task types, including attention-mediated copying versus MLP-mediated binding. ILP signatures achieve substantially better structural separation than graph kernel and feature-vector baselines, and support principled transfer across model scales and architecture families.\footnote{Code and supplementary materials are available at \url{[anonymised for review]}.}
\end{abstract}
\section{Introduction} \label{sec:introduction} 
Mechanistic interpretability (MI) aims to explain large language models (LLMs) by reverse-engineering their internal computation into human-understandable mechanisms~\citep{ferrando2024primerinnerworkingstransformerbased}. As MI scales across models, training regimes, and tasks a fundamental scientific challenge emerges: \emph{How can we transform mechanistic discoveries into cumulative knowledge that coheres across experiments, models and tasks?}~\citep{naveed2025comprehensive,zamfirescu2025beyond,ferrando2024primerinnerworkingstransformerbased}. A core focus of MI is circuit analysis, which aims to identify \emph{circuits}: sparse, causally-relevant subgraphs whose interventions reliably alter model behaviour~\citep{nanda2023progress,meng2022locating, aljaafari2025emergencelocalisationsemanticrole}. However, as more circuits are discovered, the field faces a growing problem of \emph{experimental fragmentation}: results arise from different datasets, model variants, and evaluation protocols, and are often summarised with informal labels (e.g., \emph{induction head}). \citet{merullo2024circuit} demonstrate that circuit component reuse occurs across tasks, but without a formal representation language, such reuse can only be observed post hoc and cannot be predicted or transferred systematically.
\paragraph{Problem: coherence, not just discovery.} Existing discovery pipelines, activation patching~\citep{meng2022locating}, and related attribution-based approaches~\citep{hanna2024have}, can identify causally relevant subgraphs, but they do not produce representations of \emph{what was found} that support systematic comparison and accumulation. Across experimental settings, we lack:  
\begin{figure*}[ht]\centering 
    \includegraphics[width=0.85\linewidth]{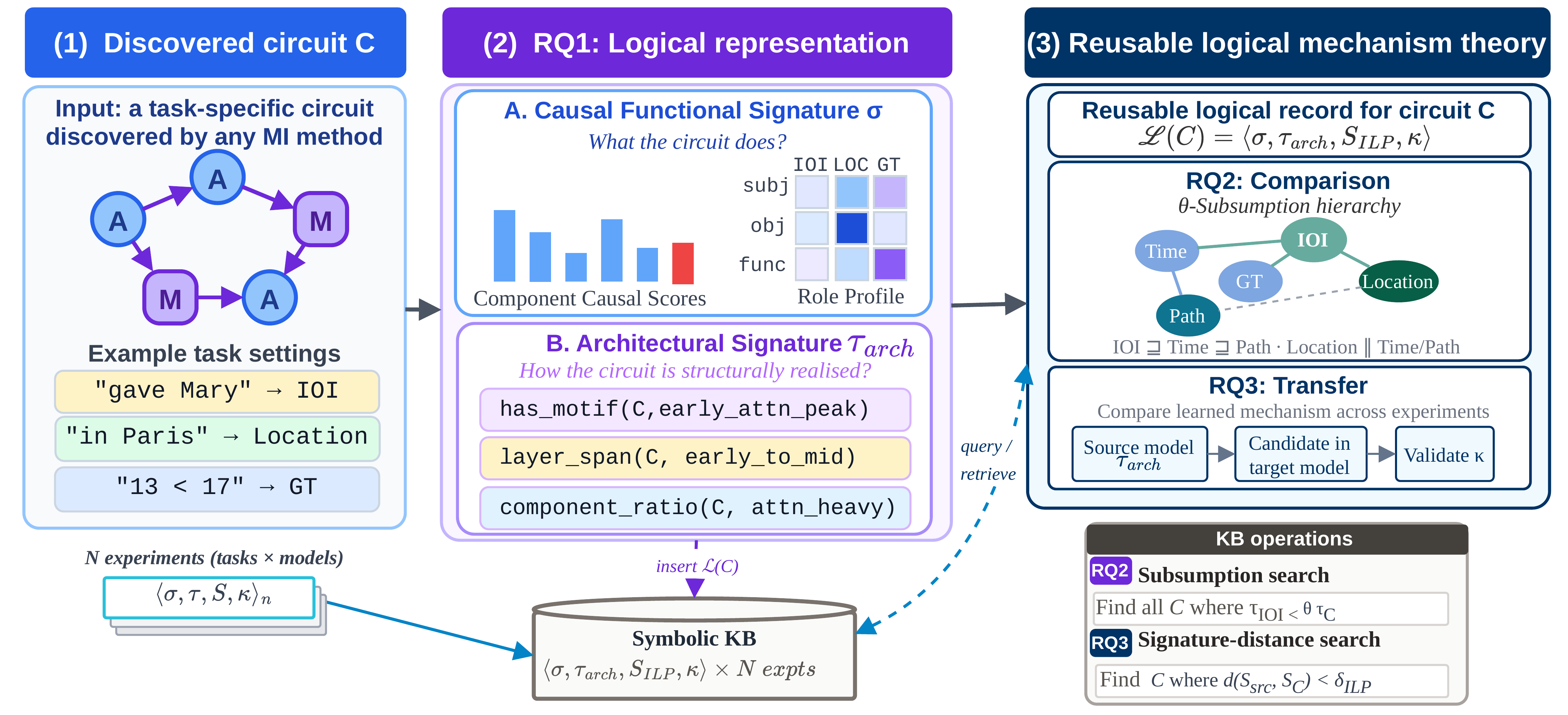} 
    \caption{\textbf{Overview for inductive circuit theory construction.} A discovered circuit~$C$ receives two complementary descriptions: a \emph{Causal Functional Signature}~$\sigma$ capturing what the circuit computes via causal attribution evidence, and an \emph{architectural signature}~$\tau_{\mathrm{arch}}$ capturing how it is structurally realised as a scale-invariant Horn clause learned by ILP\textsubscript{arch}. They form a reusable logical record $\mathcal{L}(C) = \langle\sigma,\,\tau_{\mathrm{arch}},\,\mathcal{S}_{\mathrm{ILP}},\,\kappa\rangle$ enabling formal comparison via $\theta$-subsumption (RQ2) and transfer to new models (RQ3).}
\label{fig:methodology} 
\end{figure*}
\begin{itemize}[leftmargin=*,itemsep=2pt] 
    \item \textbf{A semantic language} to state what a circuit computes beyond model-specific descriptions; 
    \item \textbf{Equivalence and refinement criteria} for deciding when new evidence confirms, specialises, or contradicts an existing mechanistic claim; 
    \item \textbf{Transfer principles} that reuse prior mechanistic hypotheses to guide new interventions across model scales. 
\end{itemize} 
As a result, the central bottleneck is formal coherence: representing mechanistic claims so they can be compared, refined, and reused across settings.
\paragraph{Approach.} We reframe circuit interpretation as \textbf{inductive theory construction} and introduce a formal coherence layer that makes MI results explicit, comparable, and reusable. From experimental traces and circuit structure, we learn \emph{logical mechanism theories} that are testable, comparable, and refinable under new evidence. Each circuit receives two coupled descriptions: 
\begin{itemize}[leftmargin=*,itemsep=1pt]
    \item \textbf{Causal Functional Signature} (\emph{what} the circuit computes), derived from causal attribution evidence over token positions labelled at two levels: shared linguistic roles enabling cross-task comparison, and task-specific roles capturing within-task functional structure;  
    \item \textbf{Architectural signature} $\tau_{\mathrm{arch}}$ (\emph{how} the mechanism is structurally realised), learned by ILP from circuit graph patterns and model-size-invariant structural features.  
\end{itemize} 
The what/how distinction is load-bearing: CFS grounds claims in causal behaviour, while $\tau_{\mathrm{arch}}$ provides portability across model scales within an architectural family. Logical representations support this by separating intended computation from incidental details, enabling structural comparison via relations such as $\theta$-subsumption, and allowing learned predicates to serve as building blocks in new hypotheses. ILP serves as the inductive engine, integrating structural motifs, attribution scores, and behavioural traces into coherent theories~\citep{muggleton1991inductive}. These capabilities motivate three research questions:
\begin{itemize}[leftmargin=*,itemsep=2pt] 
    \item \textbf{RQ1 (Representation):} To what extent can logical representations jointly capture causal behavioural evidence and structural circuit patterns in a form portable across experiments?
    \item \textbf{RQ2 (Comparison):} How well do learned logical mechanism theories support formal comparison and refinement across MI experiments? 
    \item \textbf{RQ3 (Transfer):} To what extent does prior mechanistic knowledge, captured as logical architectural signatures grounded in causal behaviour, support transfer to new models, and what is lost in the abstraction? 
\end{itemize} 
This work makes four contributions: \textbf{(i) Coherence-first formulation:} we frame circuit analysis as an inductive science of mechanisms, where the bottleneck is cross-experiment comparability and knowledge accumulation; \textbf{(ii) Causal Functional Signature (CFS):} a two-level causal characterisation of circuit behaviour using linguistic roles for cross-task comparison and task-specific roles for within-task structure; \textbf{(iii) ILP architectural signatures:} explicit logical mechanism descriptions that support testing, $\theta$-subsumption-based comparison, and cross-scale transfer; and \textbf{(iv) Evaluation:} 15 circuits ($\text{5 tasks }{\times} 3$ models), scaled to 30 circuits per model across 10 tasks and three prompt splits, with ILP signatures achieving a $3{-}4{\times}$ structural separation advantage over Weisfeiler-Lehman (WL) kernel~\citep{shervashidze2011weisfeiler} and random-forest~\citep{breiman2001random} baselines.

\section{Related work}

\paragraph{Circuit discovery and circuit-level mechanisms.}
MI studies model behaviour, including identifying circuits: subgraphs or feature pathways that causally support a task. Foundational work on transformer circuits and induction heads established circuit-level analysis as a way to relate internal components to algorithmic behaviour \citep{elhage2021mathematical,olsson2022context}. Subsequent studies have identified mechanisms for indirect object identification \citep{wang2023interpretability}, arithmetic reasoning \citep{Stolfo2023AMI}, semantic roles \citep{aljaafari2025emergencelocalisationsemanticrole}, and other task-specific behaviours. Activation patching \citep{meng2022locating}, path patching \citep{goldowsky2023localizing}, automated circuit discovery \citep{conmy2023towards}, and EAP--IG \citep{hanna2024have} provide complementary methods for identifying causally relevant components or edges. Work on circuit component reuse further suggests that mechanisms may generalise across tasks \citep{merullo2024circuit}. Our work is orthogonal to discovery: given a discovered circuit, we provide a logical specification layer for representing its causal-functional evidence, learning its architectural signature, and comparing it with other mechanism hypotheses.

\paragraph{Feature-level circuits and reusable structure.}
Sparse autoencoders decompose activations into more interpretable features \citep{huben2024sparse}; sparse feature circuits and circuit tracing extend this view by identifying causal feature-level subgraphs and attribution graphs \citep{marks2025sparse,ameisen2025circuit}. These approaches refine the granularity of circuit nodes. Our representation is agnostic to this granularity: the same logical layer can represent circuits over attention heads, MLP blocks, or finer-grained feature nodes, provided that graph structure and causal-functional evidence are available.

\paragraph{Abstraction, formality, and global perspectives.}
Causal abstraction formalises interpretability as the search for high-level causal models that preserve low-level intervention behaviour \citep{JMLR:v26:23-0058}, while causal scrubbing provides an operational framework for testing mechanistic hypotheses through intervention-based resampling \citep{chan2022causal}. Recent formal MI work applies neural-network verification to circuit discovery, seeking guarantees for robustness, patching, and minimality \citep{hadad2026formal}. At a broader level, \citet{he2025modcirc} propose global-level MI through reusable modular circuits, and cognitive studies suggest that LLMs may develop language-independent abstractions beyond surface cues \citep{chen2025emergence}. These works share our motivation to move beyond isolated circuit findings. We differ by formalising the representation layer itself: mechanism hypotheses are stored as typed logical objects that support comparison, refinement, and transfer.

\paragraph{Logic and neuro-symbolic interpretability.}
Several mechanistic studies analyse how language models implement logical or formal reasoning, including propositional logic circuits \citep{hong2024implies} and content-independent syllogistic reasoning circuits \citep{kim2025reasoning}. Related work also compares circuits for formal and functional linguistic abilities \citep{hanna2025dissociated}. \citet{palumbo2025validating} formalise mechanistic interpretations axiomatically, characterising when mechanistic descriptions approximately preserve network semantics. Our use of logic is different: rather than studying circuits that compute logical operations, we use inductive logic programming~\citep{muggleton1991inductive,cropper2022inductive} to construct architectural signatures over discovered circuits, making mechanism claims explicit, comparable, refinable, and transferable.
\section{Methodology}\label{sec:methodology}
We formalise circuit interpretation as inductive theory construction (Fig.~\ref{fig:methodology}). We introduce the core representation and learning components below, with additional ILP and architectural-signature details in Appendices~\ref{app:ilp-primer} and~\ref{app:arch-signatures}.

\subsection{Problem Formulation}
Let $M$ denote a transformer with computational graph $G=(V,E)$, and let $C=(V_C,E_C)$ be a sparse subgraph identified by any circuit discovery method. Given $C$, a task-specific evaluation suite $\mathcal{D}_T$, and background knowledge $\mathcal{B}$, we associate each circuit with a logical identity
\[
    \mathcal{L}(C){=}\langle \sigma,\,\tau_{\text{arch}},\,\mathcal{S}_{\text{ILP}},\,\kappa \rangle,
\]
where $\sigma$ is the CFS, $\tau_{\text{arch}}$ is the learned architectural signature, $\mathcal{S}_{\text{ILP}}$ summarises the learned theory, and $\kappa$ records validation statistics. All components are derived automatically from the stored circuit representation.

\subsection{Formal Circuit Representation}\label{sec:fcr}
Each circuit $C{=}(V_C,E_C)$ is represented as a set of typed ground facts organised into four layers. Layer~0 records provenance, such as task, model, and discovery method. Layer~1 records graph structure through node, type, layer, edge, component-ratio, relative-size, layer-span, and named-motif predicates, e.g., \texttt{has\_motif}$(C,m)$ and normalised faithfulness $\mathrm{Faith}(C){=}(\mathrm{Acc}_T(C){-}\mathrm{Acc}_T^{\mathrm{abl}})/(\mathrm{Acc}_T(M){-}\mathrm{Acc}_T^{\mathrm{abl}})$ \citep{mueller2025mib}. Layer~2 stores the CFS $\sigma$, and Layer~3 stores the learned architectural signature $\tau_{\mathrm{arch}}$. This representation exposes circuit evidence to ILP$_{\mathrm{arch}}$ as Prolog-style facts while keeping storage and extraction details separate (detailed extraction procedure is in App.~\ref{app:owl-foundation}).


\subsection{Causal Functional Signature (CFS)} \label{sec:CFS}
The CFS grounds each circuit component in causal behavioural evidence through two quantities. First, we compute the Direct Logit Attribution (DLA) score  \citep{nostalgebraist2020interpreting} $\delta_v {\in} \mathbb{R}$ for each node $v {\in} V_C$, measuring its additive contribution to the correct-answer logit. Components with $|\delta_v| < \epsilon_{\text{DLA}}$ are flagged as causally marginal. Second, for each causally relevant component, we compute an \emph{attribution-weighted attention profile} $\pi_v$: the attention distribution over input positions, weighted by $|\delta_v|$, with positions labelled by linguistic roles (subject, object, verb, function, entity, other) and task-specific roles (e.g., scaffold, repeated\_name; see App.~\ref{app:role-labelling}). The CFS is the collection $\sigma{=}\{\langle v, \delta_v, \pi_v^{\text{ling}}, \pi_v^{\text{task}} \rangle : v {\in} V_C\}$, stored as Layer~2 facts that enrich the background knowledge for ILP$_{\text{arch}}$.

\subsection{Learning Architectural Signatures} 
\label{sec:ilparch} 
In ILP, learning is formulated as rule induction from examples and background knowledge. In our setting, ILP$_{\mathrm{arch}}$ learns an \emph{architectural signature} $\tau_{\mathrm{arch}}$ for a circuit family. Positive examples $E^+$ are circuits that implement the target mechanism (e.g., IOI or Location binding) and negative examples $E^-$ are circuits implementing other mechanisms. The background knowledge $\mathcal{B}_{\mathrm{arch}}$ contains logical facts describing each circuit's structure. ILP then searches for a rule that covers the positive circuits while excluding the negatives. The learned rule is a Horn clause, e.g.,
\headpred{arch\_task(C)} \texttt{:-}
\bodypred{pred\_1(C)}, \bodypred{pred\_2(C,X)}, \texttt{...},
which reads as: circuit $C$ has architectural type \headpred{task} if the predicates in the body hold, making the learned signature an explicit structural hypothesis.

The predicates in $\mathcal{B}_{\mathrm{arch}}$ are chosen to be scale-invariant such that signatures can be compared across models of different scales. They include normalised depth $\rho(\ell){=}\ell/L$, component composition (\texttt{component\_ratio}, \texttt{rel\_size}), named structural motifs (\texttt{has\_motif}), and CFS facts. For example, a semantic-role circuit may be represented by facts stating that it contains an early attention peak, has a high MLP ratio, spans a particular layer range, and contains components whose CFS profiles attend to role-bearing tokens. ILP$_{\mathrm{arch}}$ uses these facts to induce a compact clause that distinguishes this circuit family from alternatives. Motifs are ranked before clause search by an attribution-weighted information gain score:
\[
\textsc{attr\_ig}(m){=}\mathrm{IG}(m) \times (1 + \bar{s}_m),
\]
where $\bar{s}_m$ is the mean edge attribution score of motif $m$ in the positive circuits. This favours motifs that are both discriminative and causally salient. We then search two complementary hypothesis spaces. Pool~A contains clauses built from named motif predicates, such as \texttt{has\_motif(C, early\_attn\_peak)}, and producing directly more interpretable signatures. Pool~B contains clauses built from blind structural predicates, such as edge-type patterns, layer-span predicates, and component-composition predicates. Pool~B acts as a check that the learned signature is not limited to the hand-defined motif names. Clause selection is based on one-vs-rest F1 over the positive and negative circuit sets: among candidate clauses from both pools, ILP$_{\mathrm{arch}}$ selects the highest-scoring clause, with shorter clauses preferred under ties.
The output is the learned Horn clause $\tau_{\text{arch}}$ with an ILP signature
\[
\mathcal{S}_{\text{ILP}}(\tau_{\text{arch}}) =
\langle n_{\text{clauses}}, \bar{d}, \mathcal{P}_H, \bar{\ell}, \texttt{complexity} \rangle,
\]
which summarises the number of clauses, derivation depth, predicate vocabulary, clause length, and composite complexity score. The clause $\tau_{\mathrm{arch}}$ serves as the interpretable mechanistic hypothesis: a logical claim that can be read, challenged, and refined. The summary $\mathcal{S}_{\mathrm{ILP}}$ serves as its compact computational handle, enabling distance-based comparison between signatures and candidate matching during transfer. Full definitions of motif extraction, motif scoring, dual-pool search, and ILP signature construction are provided in App.~\ref{app:arch-signatures}.
\subsection{Validation Criteria}
\label{sec:validation}
A circuit $C$ is retained as a valid mechanism candidate only if it satisfies behavioural consistency and causal relevance. Behavioural consistency requires the circuit to support the target behaviour on the evaluation suite, $\mathrm{Acc}_T(C)\geq\theta_{\mathrm{behav}}$. Causal relevance measures the task-performance degradation caused by ablating the circuit:
\[
    \Delta_T(C)=\mathrm{Acc}_T(M)-
    \mathrm{Acc}_T(M^{\mathrm{ablate},C}).
\]
We require $\Delta_T(C)\geq\theta_{\mathrm{causal}}$ and check approximate $\epsilon$-minimality by greedy edge ablation. Threshold values, random-subgraph significance tests, and sensitivity analyses are in App.~\ref{app:hyperparameters}.

\subsection{Comparison via $\theta$-Subsumption}
\label{sec:comparison}
Learned clauses support cross-experiment comparison via $\theta$-subsumption~\citep{gottlob1987subsumption}. A clause $\gamma_1$ subsumes a clause $\gamma_2$, written $\gamma_1 \preceq_\theta \gamma_2$, when some substitution $\theta$ makes every literal in $\gamma_1\theta$ appear in $\gamma_2$. This induces an ordering over learned signatures: one signature may generalise or specialise another, two signatures may be equivalent under mutual subsumption, or they may be incomparable. In the relaxed comparisons, numeric thresholds are abstracted away, so the ordering captures predicate-level structure rather than exact threshold agreement. We use this relation to identify when one mechanism theory makes stronger structural commitments than another; strict and relaxed definitions are given in App.~\ref{app:subsumption}.

\subsection{Transfer of Mechanistic Knowledge}
\label{sec:transfer}
Given a source mechanism record $\mathcal{L}(C_\alpha)$ in model $M_\alpha$ and a target model $M_\beta$, transfer treats the source architectural signature as a retrieval hypothesis. The procedure has four stages: (i) retrieve $\tau_{\mathrm{arch},\alpha}$ and $\sigma_\alpha$ from the registry; (ii) identify candidate circuits in $M_\beta$ whose scale-normalised structural predicates and ILP signature distance match the source signature within tolerance; (iii) screen candidates using behavioural and causal validation on the target evaluation suite; and (iv) select the best validated candidate. If no candidate passes validation, we fall back to full circuit discovery in the target model and compare the re-learned signature against the source signature. Thus, transfer is not treated as proof of mechanism identity from structure alone; it is a proposal-and-validation procedure in which logical abstraction retrieves candidates and empirical tests determine whether the transfer is supported. The full transfer algorithm, including tolerance values and fallback criteria, is given in App.~\ref{app:algorithms}.
\subsection{Experimental Setup}\label{sec:experiments}
We evaluate on three task families covering distinct operation types: semantic role binding ($\kappa{=}\texttt{binding}$), indirect object identification ($\kappa{=}\texttt{selection}$), and numerical comparison ($\kappa{=}\texttt{comparison}$). Circuits are extracted using EAP-IG~\citep{hanna2024have} with $k{=}200$ edges retained per circuit; the framework accepts any discovery method producing attributed subgraphs (Section~\ref{sec:fcr}). The primary evaluation covers 15 circuits (5 tasks $\times$ 3 models): semantic roles Location, Time, and Path~\citep{aljaafari2025emergencelocalisationsemanticrole}, IOI~\citep{wang2023interpretability}, and Greater-Than, across Pythia (14M and 1B)~\citep{biderman2023pythia}, and LLaMA-3.2-1B~\citep{grattafiori2024llama}. The scaled evaluation expands to 10 task types (8 semantic roles, IOI, Greater-Than) with 3 circuits per task discovered from disjoint prompt subsets, yielding 30 circuits per model. Task selection rationale, model cards, and implementation details are in Appendix~\ref{app:task-model-selection} and~\ref{app:implementation}. Full circuit inventory (node counts and split) is in Appendix~\ref{app:circuit-inventory}.

\section{Results}\label{sec:results}
We report the main findings for each RQ below; supporting tables and extended results are provided in App.~\ref{app:additional-results}.
\subsection{CFS Reveals Distinct Computational Strategies}
\label{sec:rq1-results}
\begin{table}[t]
\centering
\small
\resizebox{\linewidth}{!}{
\begin{tabular}{@{}llrrcr@{}}
\toprule
\textbf{Model} & \textbf{Task} & \textbf{Caus.} & \textbf{Marg.} & \textbf{Dom.\ Role} & \textbf{Faith.} \\
\midrule
\multirow{5}{*}{Py-14M} & IOI & 26 & 0 & object (50\%) & 0.79 \\
& Location & 21 & 1 & object (76\%) & 0.57 \\
& Time & 21 & 0 & function (81\%) & 1.22 \\
& Path & 12 & 12 & object (71\%) & 1.21 \\
& GT & 28 & 0 & entity (77\%) & 0.88 \\
\midrule
\multirow{5}{*}{Py-1B} & IOI & 34 & 2 & function (58\%) & 0.14 \\
& Location & 29 & 3 & object (74\%) & 0.76 \\
& Time & 28 & 2 & function (87\%) & $-$7.23\textsuperscript{\dag} \\
& Path & 31 & 5 & object (55\%) & 1.08 \\
& GT & 40 & 0 & function\textsuperscript{$\star$} (48\%) & 0.85 \\
\midrule
\multirow{5}{*}{LL-1B} & IOI & 0 & 29 & function (67\%)\textsuperscript{\ddag} & 0.01 \\
& Location & 8 & 23 & object (71\%) & 0.32 \\
& Time & 5 & 21 & function (100\%)\textsuperscript{\S} & 0.88 \\
& Path & 10 & 18 & object (68\%) & 0.18 \\
& GT & 11 & 18 & function (58\%) & 0.05 \\
\bottomrule
\end{tabular}
}
\caption{\textbf{CFS summary across models.} Dom.\ Role = most common $\arg\max \pi_v$ over linguistic roles.  \dag~Near-marginal circuit ($\Delta_T=0.003$).  \ddag~All LLaMA-IOI components are causally marginal despite $\Delta_T=5.24$. $\star$~Tie between \texttt{function} and \texttt{entity}. \S~All active components attend to function words.}
\label{tab:cfs-summary}
\vspace{-1.2em}
\end{table}

\paragraph{Role-consistent attention across scales.} CFS profiles (Tab.~\ref{tab:cfs-summary}) show that dominant attention roles are determined more by task type than scale. Location and Path circuits are \texttt{object}-dominant at every scale (Location: 76\%, 74\%, 71\%; Path: 71\%, 55\%, 68\% across Pythia-14M, Pythia-1B, LLaMA), consistent with circuits attending to the grammatical object position hosting the role filler. Time circuits are \texttt{function}-dominant throughout (81\%, 87\%, 100\%\textsuperscript{\S}), reflecting attention to temporal prepositions and determiners. GT circuits are \texttt{entity}-dominant at 14M (77\%) but shift toward \texttt{function}-dominant at larger scales (48\%\textsuperscript{$\star$}, 58\%). IOI shifts from \texttt{object}-dominant at 14M (50\%) to \texttt{function}-dominant at 1B and LLaMA (58\%, 67\%), suggesting that the copy mechanism relocates attention from name positions to structural tokens as capacity increases. CFS feature vectors cluster significantly by task family on Pythia-1B ($p{=}0.008$, permutation test). LLaMA supports task-type consistency across the evaluated architectural families. The LLaMA IOI circuit has all 29 components causally marginal ($|\delta_v|{<}0.05$) despite high circuit-level causal relevance ($\Delta_T{=}5.24$); its full-component profile is nonetheless \texttt{function}-dominant (67\%), matching the Pythia-1B pattern and suggesting a distributed implementation of the same functional strategy.

\begin{figure}[t]
\centering
\includegraphics[width=0.8\linewidth]{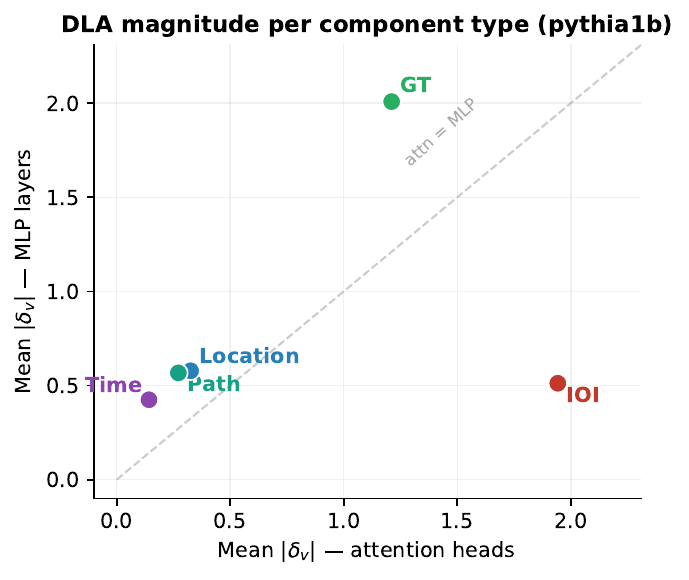}
\caption{\textbf{DLA magnitude per component type (Pythia-1B).} Mean $|\delta_v|$ for attention heads (x-axis) vs.\ MLP layers (y-axis). IOI falls \emph{below} the diagonal (attn-dominant, $|\delta_v|_{\mathrm{attn}} \approx 2.0$); Location, Time, and Path fall \emph{above} (MLP-dominant). GT is MLP-dominant with high absolute magnitude ($|\delta_v|_{\mathrm{MLP}} \approx 2.0$), consistent with MLP-mediated numerical comparison.} 
\label{fig:dla}
\vspace{-0.5em}
\end{figure}

\noindent\textbf{Distinct wiring: attention copying vs.\ MLP binding.} The CFS differences reflect qualitatively different circuit wiring (App.~\ref{app:edge-types}). In Pythia-1B, IOI routes information through 85 attn$\to$attn edges, forming three-hop attention chains (\texttt{attn\_chain\_3}) characteristic of the copy mechanism. Semantic role circuits rely instead on 64--96 mlp$\to$mlp edges, with motifs such as \texttt{mlp\_heavy} reflecting MLP-dominated information flow; LLaMA preserves this pattern (44 vs.\ 0--2 attn$\to$attn edges). Fig.~\ref{fig:dla} provides a complementary attribution-level view: IOI falls \emph{below} the diagonal (attn-dominant), while Location, Time, and Path fall \emph{above} (MLP-dominant). GT also falls above the diagonal but with high absolute MLP magnitude, consistent with MLP-mediated numerical comparison rather than distributed binding. Full linguistic role profiles are in App.~\ref{app:role-labelling}.
\vspace{-0.3em}
\subsection{ILP Captures Structural Boundaries}
\label{sec:rq2-results}

\paragraph{Learned clauses encode the binding/selection distinction.}
ILP$_{\text{arch}}$ learns one Horn clause per task via one-vs-rest classification (Tab.~\ref{tab:arch-example}; full results in App.~\ref{app:learned-clauses}). In the 5-task pool, the IOI clause requires only a single predicate, high attention ratio, directly reflecting the attention-dominated wiring identified by CFS; in the expanded 10-task pool, IOI acquires additional predicates (\texttt{attn\_chain\_3}, \texttt{layer\_span}, \texttt{size}) to maintain discrimination against structurally similar roles (full clauses in App.~\ref{app:learned-clauses}). Role and GT clauses require conjunctions of motifs, component ratios, and size constraints in both pools, capturing the MLP-heavy composition that no single feature can express. Thus, the clause structure turns the mechanistic distinction into a testable structural hypothesis: circuits satisfying the IOI clause are retrieved as candidates for attention-mediated copying, and CFS with $\Delta_T$ determines whether that interpretation is empirically supported.
\begin{table}[t]
\centering
\scriptsize
\setlength{\tabcolsep}{2.5pt}
\renewcommand{\arraystretch}{0.95}
\setlength{\fboxsep}{1.1pt}
\setlength{\fboxrule}{0pt}
\begin{tabular}{@{}p{0.72cm}p{\dimexpr\linewidth-1.52cm\relax}c@{}}
\toprule
\textbf{Task} & \textbf{Predicates in $\tau_{\text{arch}}$} & $\mathcal{C}$ \\
\midrule
IOI
&
\pcomp{\texttt{comp\_ratio(C,\,attn,\,R)}} \;
\texttt{R > 0.63}
& 0.2 \\[2pt]
LOC
&
\pmotif{\texttt{has\_motif(C,\,mlp\_heavy)}} \;
\pmotif{\texttt{has\_motif(C,\,mid\_layer\_attn\_peak)}} \;
\pspan{\texttt{layer\_span(C,\,L,\,L')}} \;
\pcomp{\texttt{comp\_ratio(C,\,attn,\,R)      }                                    } \;     
\texttt{R > 0.38} \;
\pcomp{\texttt{size(C,\,N)}} \;
\texttt{N{<}38}
& 0.9 \\
\bottomrule
\end{tabular}
\caption[\textbf{Example architectural signatures (Pythia-1B, 5-task pool).}]{ \textbf{Example architectural signatures (Pythia-1B, 5-task pool).} IOI is structurally simple, while Location requires a conjunctive signature. Colour identifies predicate type: \protect\fcolorbox{motiffg}{motifbg}{\protect\textcolor{motiffg}{\texttt{named structural motif}}}; \protect\fcolorbox{spanfg}{spanbg}{\protect\textcolor{spanfg}{\texttt{layer position}}}; \protect\fcolorbox{compfg}{compbg}{\protect\textcolor{compfg}{\texttt{circuit composition}}}.}
\label{tab:arch-example}
\vspace{-1.2em}
\end{table}

\begin{figure}[t]
\centering
\centering
\includegraphics[width=0.75\linewidth]{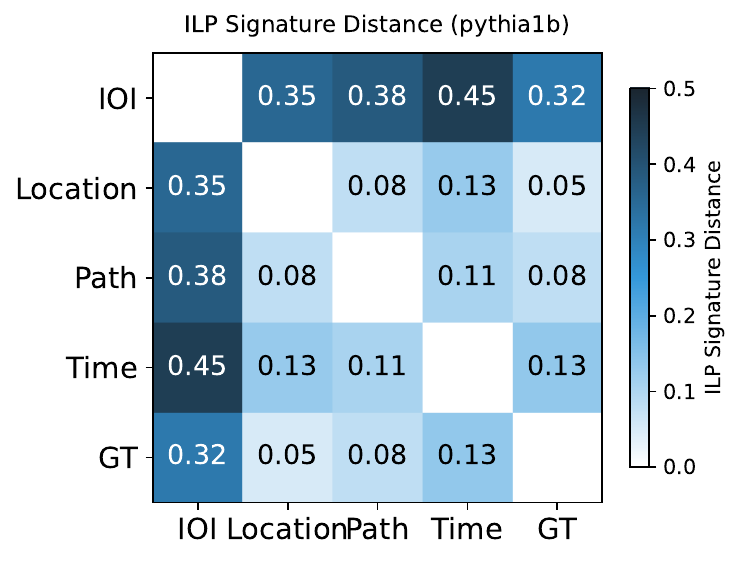}
\caption{\textbf{ILP signature distance (Pythia-1B).} IOI is distant from all circuits ($d \geq 0.32$). Roles and GT cluster tightly ($d \leq 0.13$).}
\label{fig:distance}
\end{figure}

\begin{figure}[t]
\centering
\includegraphics[width=0.8\linewidth]{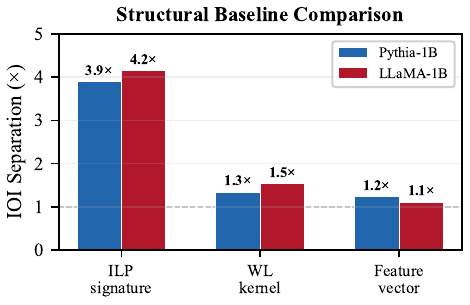}
\caption{\textbf{IOI separation: ILP vs.\ structural baselines.} Grouped by model. ILP achieves $3.9$--$4.2\times$; WL kernel $1.3$--$1.5\times$.}
\label{fig:baselines}
\vspace{-1.1em}
\end{figure}

\noindent\textbf{ILP distance reveals task family structure.}
Fig.~\ref{fig:distance} shows the pairwise ILP signature distance on Pythia-1B. IOI is distant from all other circuits ($d{=}0.32$--$0.45$), while semantic roles and GT form a tight cluster ($d{=}0.05$-$0.13$). GT clusters with roles because its clause uses the same predicate vocabulary despite implementing a different computation, suggesting that GT circuits are realised through structural patterns closer to binding than to attention-copying. The threshold $\delta_{\text{ILP}}{=}0.30$ cleanly separates IOI from all other circuits; the same pattern holds on LLaMA ($d_{\text{IOI}}{=}0.34$--$0.47$; see App.~\ref{app:full-distance} for all models). We compare ILP distance against three structural alternatives (Fig.~\ref{fig:baselines}). These are a Weisfeiler-Lehman (WL) graph kernel~\citep{shervashidze2011weisfeiler} capturing local neighbourhood structure, a random forest~\citep{breiman2001random} trained on the same structural features used as ILP background knowledge, and Euclidean distance over the same feature vectors. ILP achieves $3.9\times$ IOI separation on Pythia-1B ($4.2\times$ on LLaMA), compared to $1.3\times$ for the WL kernel. The random forest has access to the same predicates, but does not return compact inspectable clauses over those predicates; its 60\% leave-one-out accuracy (misclassifying GT and IOI) indicates that explicit logical structure provides useful discrimination beyond the feature inventory alone (details in App.~\ref{app:baseline-details}).

\noindent\textbf{Subsumption reveals a non-trivial refinement hierarchy.}
Using the 5-task-pool clauses under relaxed $\theta$-subsumption, where numeric thresholds are abstracted away and predicate structure determines the ordering, a non-trivial hierarchy emerges (Tab.~\ref{tab:subsumption}). In this relaxed sense, the single-predicate IOI clause generalises the other clause, any circuit satisfying a role or GT clause also satisfies the attention-ratio constraint. Within the role family, Time generalises Path, which adds \texttt{attn\_chain\_3}, while Location is incomparable to both because it uses different motifs (\texttt{mlp\_heavy}, \texttt{mid\_layer\_attn\_peak}). The hierarchy also changes with scale: in Pythia-14M, Time's size-only clause generalises all others and Location $\equiv$ Path, reflecting greater structural homogeneity; in Pythia-1B, the richer hierarchy emerges because the larger model produces circuits diverse enough to support meaningful differentiation. This partial order formalises relationships that informal labels and distance metrics do not express: Location and Time are not simply close or far, but distinct structural specialisations of a broader binding template (App.~\ref{app:subsumption}).

\noindent\textbf{Scaled evaluation and within-task consistency.}
The expanded 10-task evaluation shows that ILP$_{\text{arch}}$ remains informative beyond the 5-task pool, but also exposes genuine within-task variability. Structurally stable tasks retain high confidence (IOI: 0.93, GT: 0.89), whereas prompt-sensitive semantic roles drop substantially (Location: 0.31, Beneficiary: 0.31). Extended distance matrices, within-task variance, and sensitivity analyses are reported in Appendices~\ref{app:full-distance}--\ref{app:sensitivity}.
\begin{table}[t]
\centering
\small
\captionsetup{justification=raggedright,singlelinecheck=false}
\begin{tabular}{@{}lccccc@{}}
\toprule
& \textbf{IOI} & \textbf{LOC} & \textbf{PATH} & \textbf{TIME} & \textbf{GT} \\
\midrule
\textbf{IOI}  & $=$ & $\sqsupseteq$ & $\sqsupseteq$ & $\sqsupseteq$ & $\sqsupseteq$ \\
\textbf{LOC}  & $\sqsubseteq$ & $=$ & $\|$ & $\|$ & $\|$ \\
\textbf{PATH} & $\sqsubseteq$ & $\|$ & $=$ & $\sqsubseteq$ & $\|$ \\
\textbf{TIME} & $\sqsubseteq$ & $\|$ & $\sqsupseteq$ & $=$ & $\|$ \\
\textbf{GT}   & $\sqsubseteq$ & $\|$ & $\|$ & $\|$ & $=$ \\
\bottomrule
\end{tabular}
\caption{\textbf{Relaxed $\theta$-subsumption (Pythia-1B).} $\sqsupseteq${=}generalises; $\sqsubseteq${=}specialises; $\|${=}incomparable. IOI generalises all; Time generalises Path; Location is incomparable to Time/Path.}
\label{tab:subsumption}
\vspace{-1.1em}
\end{table}
\subsection{ILP Signatures Support Transfer}
\label{sec:rq3-results}

\textbf{Scale-invariant predicates enable cross-scale and cross-family transfer.}
Raw component overlap is near zero across all model pairs (node Jaccard $0.09$--$0.24$ within Pythia; lower cross-family), thus transfer via node identity is not possible. Scale-invariant predicates (\texttt{component\_ratio}, \texttt{layer\_span}, \texttt{has\_motif}) abstract over these differences. On Pythia-1B each semantic role circuit finds 2--3 candidates within $\delta_{\text{ILP}} = 0.30$, while IOI finds none ($d = 0.32$--$0.45$). LLaMA shows the same separation pattern ($d_{\text{IOI}}{=}0.34$--$0.47$), confirming that the boundary between attention-copying and MLP-binding mechanisms is architecture-independent.

\noindent\textbf{Live transfer across scales and families.}
End-to-end transfer accepts candidates for all 5 tasks in both directions: Pythia-14M$\to$Pythia-1B (cross-scale, $70\times$ gap) and LLaMA$\to$Pythia-1B (cross-family). For IOI and GT, the best accepted candidate matches the same task in the target model, with high selectivity (2/37--41 for GT, 8/41 for IOI). For semantic roles, the best candidate is Goal, a role not in the source pool, because it has the highest $\Delta_T$ among binding circuits in Pythia-1B ($\Delta_T{=}1.00$). This is a semantically valid match: both implement binding via the same motif vocabulary. The accept/reject pattern is identical across both transfer directions despite very different source $\Delta_T$ values (e.g.\ IOI: $-0.91$ from Pythia-14M vs.\ $5.24$ from LLaMA), confirming that the structural signature, not the source behaviour, drives candidate selection (App.~\ref{app:live-transfer}).
\noindent\textbf{Binding mechanisms preserve causal contribution; selection and comparison do not.}
Tab.~\ref{tab:cross-scale-dt} reports $\Delta_T$ across all three models. In this evaluation, causal preservation aligns more closely with mechanism type than with architectural similarity. Within the Pythia family, IOI undergoes a sign reversal from 14M to 1B ($-0.91{\to}4.37$), indicating a qualitative change in the functional role of the circuit with scale. GT shows a large gap across pairs (3.97 within Pythia, 2.74 cross-family). Among binding mechanisms, Location and Path show strong cross-family preservation ($|$diff$|{=}0.08$ and $0.26$ respectively), while Time does not ($|$diff$|{=}0.85$), likely because its near-zero $\Delta_T$ on Pythia-1B (0.003) makes this comparison unreliable. Where binding circuits have robust causal contributions, MLP-mediated binding appears more stable across architectures than attention-mediated copying or numerical comparison. To confirm that these causal contributions reflect structured architectural rather than chance, we evaluate 50 size-matched random circuits per task across both Pythia models (34 circuits on 14M, 35 on 1B, including multi-split variants). Random subgraphs consistently yield $\Delta_T{\approx}0$, and real circuits are far better in 29/34 cases on Pythia-14M and 28/35 on Pythia-1B ($p{<}0.05$). See App.~\ref{app:random-baseline} for per-circuit details.
\begin{table}[t]
\centering
\small
\begin{tabular}{@{}lrrrr@{}}
\toprule
\textbf{Task} & \textbf{Py-14M} & \textbf{Py-1B} & \textbf{LL-1B} & \textbf{LL$\leftrightarrow$Py-1B} \\
\midrule
IOI       & $-$0.91 & 4.37 & 5.24 & 0.87 \\
Location  & 1.93 & 1.28 & 1.20 & \textbf{0.08} \\
Time      & 0.75 & 0.003 & $-$0.85 & 0.85 \\
Path      & 0.53 & 0.21 & 0.48 & 0.26 \\
GT        & 7.95 & 3.97 & 1.23 & 2.74 \\
\bottomrule
\end{tabular}
\caption{\textbf{$\Delta_T$ across models and families.} Binding mechanisms (Location, Path) preserve causal contribution across architectures ($|$diff$|{=}0.08$--$0.26$); selection (IOI) undergoes a sign reversal across Pythia scales; comparison (GT) shows large gaps across all pairs.}
\label{tab:cross-scale-dt}
\vspace{-1.2em}
\end{table}
\section{Conclusion}
\label{sec:conclusion}
Mechanistic interpretability produces circuit discoveries that are locally valid but difficult to accumulate: without a shared representation, it remains unclear how findings from different experiments, models, or tasks should be compared. We addressed this coherence problem by treating circuit interpretation as inductive theory construction. Causal Functional Signatures ground circuit components in causal attribution evidence, while ILP-learned architectural signatures capture structural regularities as inspectable clauses. Together, they form a representation layer in which mechanistic claims can be stated explicitly, compared via $\theta$-subsumption, and reused as transfer hypotheses.
The results support three findings. First, CFS distinguishes computational strategies such as attention-mediated copying and MLP-mediated binding that are preserved across architectural families. Second, ILP signatures capture structural boundaries that raw component overlap and feature-vector baselines do not expose as clearly. Third, transfer succeeds most reliably when treated as candidate retrieval followed by behavioural and causal validation: binding mechanisms with robust causal contributions show stronger cross-architecture preservation than the selection and comparison mechanisms evaluated here. 
This work provides a formal infrastructure between circuit discovery and knowledge accumulation. Proxy tasks establish local validity; our method formalises generalisability, cross-experiment relationships, and the level of abstraction at which a claim holds. Extending the approach to more distributed mechanisms and broader architectural families remains the primary direction for future work.

\section*{Limitations}
\label{sec:limitations}
Each $\tau_{\text{arch}}$ clause is falsifiable in principle, but we test only whether it predicts causal contribution across models, not task behaviour on novel prompts, the level at which our framework and proxy-task approaches are complementary rather than competing. The evaluation covers 30 circuits across 10 task types, two Pythia scales, and one cross-family model (LLaMA-3.2-1B); the binding/selection/comparison schema is not a universal theory of internal mechanisms, and extending to additional architectural families remains important future work. All thresholds are empirically chosen; sensitivity analysis confirms stability across the primary range (Appendix~\ref{app:sensitivity}). CFS role profiles depend on a lightweight dependency parser, and the greedy $\epsilon$-minimality procedure may miss non-contiguous minimal subcircuits.

\bibliography{custom}

\appendix
\section{Brief Primer on Inductive Logic Programming}
\label{app:ilp-primer}

Inductive logic programming (ILP) \citep{muggleton1991inductive} is a subfield of machine learning that combines inductive inference with symbolic knowledge representation~\citep{cropper2022inductive}. Unlike many machine learning algorithms that produce numeric parameters (weights, probabilities), such as probes (linear classifiers over representations), graph similarity kernels (pairwise distance scores), ILP produces hypotheses expressed in first-order logic (FOL): explicit rules that can be read, inspected, compared, and transferred independently of the model that produced them.

\subsection{Positive Examples, Negative Examples, and Background Knowledge}
An ILP problem consists of positive examples $E^+$, negative examples $E^-$, and background knowledge $\mathcal{B}$. The positive examples are cases that the target concept should cover; the negative examples are cases it should exclude; and the background knowledge contains known facts and relations about the domain. ILP searches for a hypothesis $\tau$ such that:
\begin{equation}
    \mathcal{B} \cup \tau \models E^+ \quad \text{and} \quad 
    \mathcal{B} \cup \tau \not\models E^-.
\end{equation}
In practice, this means that $\tau$ should explain the positive examples while avoiding the negative examples, subject to constraints such as clause length, allowed predicates, and search depth.

\subsection{Horn Clauses}
The hypothesis $\tau$ takes the form of a \emph{Horn clause}: a rule with a single conclusion (the \emph{head}) and one or more conditions (the \emph{body}): \texttt{head(X) :- condition\_1(X), condition\_2(X, Y), ...}.
read as: ``\texttt{head} holds of \texttt{X} if all conditions hold.''

\subsection{Why ILP is Suitable for Circuit Signatures}
One could argue that simpler methods are easier to implement and may give a similar level of empirical separation. We argue that we need \emph{theory formation}: descriptions of circuit mechanisms that are explicit, formally comparable, and transferable across models. ILP satisfies these requirements by construction, providing a level of formality and interpretability that purely statistical or feature-based methods do not. In particular, ILP is useful for three reasons \citep{russell2010artificial}:
\begin{itemize}[leftmargin=*,itemsep=3pt]
    \item \textbf{Relational representation.} ILP can learn concepts that are difficult to express using fixed attribute vectors, because it represents relationships between entities rather than properties of individual ones;
    \item \textbf{Inspectable hypotheses.} ILP produces rules in a form that is directly readable: a Horn clause states explicitly \emph{why} a concept holds, not \emph{that} it holds. A human can examine the rule, challenge it, and refine it against new evidence;
    \item \textbf{Formal comparison and transfer.} As the hypotheses share the same logical language, they can be compared via $\theta$-subsumption to determine when one mechanism generalises another (Appendix~\ref{app:subsumption}), and transferred to new models without retraining.
\end{itemize}
Consider learning the concept \emph{grandparent} from family relations. The background knowledge $\mathcal{B}$ contains facts of the form \texttt{parent(X,\,Y)}. The examples are:

\begin{center}
\small
\begin{tabular}{@{}ll@{}}
\toprule
\textbf{Positive} $E^+$ & \textbf{Negative} $E^-$ \\
\midrule
\texttt{grandparent(sara, carol)} & \texttt{grandparent(sara, bob)} \\
\texttt{grandparent(bob, dave)}    & \texttt{grandparent(carol, sara)} \\
\bottomrule
\end{tabular}
\end{center}
Given these examples and the background facts, ILP can induce the clause \texttt{grandparent(X, Z) :- parent(X, Y), parent(Y, Z).}
It illustrates all three properties above. First, the concept cannot be captured by isolated attributes (age, gender) because it depends on a \emph{chain of relations} between entities, expressed here via the shared variable \texttt{Y}. Second, the rule is immediately inspectable: one can read it, verify it against new cases, and refine it by adding or removing body literals. Third, the rule can be compared against a more specific hypothesis: \texttt{grandparent(X, Z) :- parent(X, Y), parent(Y, Z), female(X).}
This variant covers only grandmothers. The original rule \emph{generalises} it: every case covered by the second is also covered by the first, but not vice versa. This relation is formalised by $\theta$-subsumption (Appendix~\ref{app:subsumption}).

\subsection{Mapping ILP to Mechanistic Interpretability}
The previous example connects naturally to mechanistic interpretability. The entities are neural network components rather than people, the background knowledge $\mathcal{B}_{\text{arch}}$ contains relational facts describing how those components are connected, and the target concept is a mechanistic type such as ``this subgraph implements location-role binding''. Positive examples are subgraphs that implement the target mechanism; negative examples are subgraphs that implement something else, for instance indirect object identification or numerical comparison. ILP induces a rule, the architectural signature $\tau_{\text{arch}}$, that captures exactly what structural properties distinguish the positive examples from the negative ones, just as the grandparent rule captures what relational property distinguishes grandparents from non-grandparents.

In both ILP and circuit analysis, complex behaviour is explained by identifying structured, interpretable components and the relations between them. In ILP, a higher-level concept such as \emph{grandparent} is built from simpler relational components, e.g., \texttt{parent(X,\,Y)} and \texttt{parent(Y,\,Z)}, linked through a shared variable. In circuit analysis, a mechanism is characterised by how its components are connected and what structural patterns they form. ILP$_{\text{arch}}$ utilises this parallel directly: rather than representing a circuit as a flat feature vector, it expresses the circuit's structural identity as a Horn clause over relational predicates, capturing dependencies that isolated attributes cannot. For example, ILP$_{\text{arch}}$ induces the following clauses for two semantic role circuits in Pythia-1B:
\begin{verbatim}

arch_time(C) :- has_motif(C, early_attn_peak),
                layer_span(C, L_min, L_max),
                component_ratio(C, attn, R), R > 0.35,
                size(C, N), N < 36.

arch_path(C) :- has_motif(C, early_attn_peak),
                has_motif(C, attn_chain_3),
                layer_span(C, L_min, L_max),
                component_ratio(C, attn, R), R > 0.44,
                size(C, N), N < 42.
\end{verbatim}
The Path clause extends the Time clause by adding one predicate, \texttt{has\_motif(C, attn\_chain\_3)}, making it structurally more specific. Under relaxed $\theta$-subsumption, Time therefore generalises Path: every circuit satisfying the Path clause also satisfies the Time clause, but not vice versa. The full refinement hierarchy is reported in Section~\ref{sec:rq2-results}. A full sample of learned clauses is given in Table~\ref{tab:arch-signatures-full}.
\begin{table*}[h]
\centering
\small
\setlength{\fboxsep}{1.2pt}
\setlength{\fboxrule}{0pt}
\setlength{\tabcolsep}{6pt}
\begin{tabular}{@{}l p{11.5cm} cc@{}}
\toprule
\textbf{Task}
  & \textbf{Learned clause $\tau_{\text{arch}}$ (Pythia-1B, 10-task pool)}
  & \textbf{Conf.} & \textbf{Cmplx.} \\
\midrule
\multicolumn{4}{@{}l@{}}{\footnotesize\textit{High confidence — single discriminative motif}} \\[2pt]

IOI
  & \texttt{arch\_ioi(C)~:-}
    \newline\hspace*{6pt}\pmotif{has\_motif(C,\,attn\_chain\_3)},
    \pspan{layer\_span(C,\,$L$,\,$L'$)},
    \pcomp{comp\_ratio(C,\,attn,\,R)}, R\,>\,0.58,
    \pcomp{size(C,\,N)}, N\,<\,44.
  & 0.93 & 0.8 \\[6pt]

GT
  & \texttt{arch\_gt(C)~:-}
    \newline\hspace*{6pt}\pmotif{has\_motif(C,\,mid\_layer\_attn\_peak)},
    \pspan{layer\_span(C,\,$L$,\,$L'$)},
    \pcomp{comp\_ratio(C,\,attn,\,R)}, R\,>\,0.50,
    \pcomp{size(C,\,N)}, N\,<\,49.
  & 0.89 & 0.8 \\[8pt]

\multicolumn{4}{@{}l@{}}{\footnotesize\textit{Intermediate confidence — two-motif conjunctions}} \\[2pt]

Source
  & \texttt{arch\_source(C)~:-}
    \newline\hspace*{6pt}\pmotif{has\_motif(C,\,attn\_chain\_3)},
    \pmotif{has\_motif(C,\,early\_attn\_peak)},
    \pspan{layer\_span(C,\,$L$,\,$L'$)},
    \pcomp{comp\_ratio(C,\,attn,\,R)}, R\,>\,0.40,
    \pcomp{size(C,\,N)}, N\,<\,42.
  & 0.81 & 0.9 \\[6pt]

Time
  & \texttt{arch\_time(C)~:-}
    \newline\hspace*{6pt}\pmotif{has\_motif(C,\,mlp\_heavy)},
    \pmotif{has\_motif(C,\,early\_attn\_peak)},
    \pspan{layer\_span(C,\,$L$,\,$L'$)},
    \pcomp{comp\_ratio(C,\,attn,\,R)}, R\,>\,0.37,
    \pcomp{size(C,\,N)}, N\,<\,37.
  & 0.81 & 0.9 \\[6pt]

Instrument
  & \texttt{arch\_instrument(C)~:-}
    \newline\hspace*{6pt}\pmotif{has\_motif(C,\,attn\_to\_attn)},
    \pmotif{has\_motif(C,\,mlp\_heavy)},
    \pspan{layer\_span(C,\,$L$,\,$L'$)},
    \pcomp{comp\_ratio(C,\,attn,\,R)}, R\,>\,0.38,
    \pcomp{size(C,\,N)}, N\,<\,39.
  & 0.52 & 0.9 \\[6pt]

Path
  & \texttt{arch\_path(C)~:-}
    \newline\hspace*{6pt}\pmotif{has\_motif(C,\,early\_attn\_peak)},
    \pmotif{has\_motif(C,\,mlp\_heavy)},
    \pspan{layer\_span(C,\,$L$,\,$L'$)},
    \pcomp{comp\_ratio(C,\,attn,\,R)}, R\,>\,0.38,
    \pcomp{size(C,\,N)}, N\,<\,39.
  & 0.52 & 0.9 \\[6pt]

Topic
  & \texttt{arch\_topic(C)~:-}
    \newline\hspace*{6pt}\pmotif{has\_motif(C,\,early\_attn\_peak)},
    \pmotif{has\_motif(C,\,attn\_to\_attn)},
    \pspan{layer\_span(C,\,$L$,\,$L'$)},
    \pcomp{comp\_ratio(C,\,attn,\,R)}, R\,>\,0.44,
    \pcomp{size(C,\,N)}, N\,<\,44.
  & 0.44 & 0.9 \\[8pt]

\multicolumn{4}{@{}l@{}}{\footnotesize\textit{Low confidence — structurally variable binding roles}} \\[2pt]

Goal
  & \texttt{arch\_goal(C)~:-}
    \newline\hspace*{6pt}\pmotif{has\_motif(C,\,early\_attn\_peak)},
    \pspan{layer\_span(C,\,$L$,\,$L'$)},
    \pcomp{comp\_ratio(C,\,attn,\,R)}, R\,>\,0.44,
    \pcomp{size(C,\,N)}, N\,<\,42.
  & 0.37 & 0.8 \\[6pt]

Beneficiary
  & \texttt{arch\_beneficiary(C)~:-}
    \newline\hspace*{6pt}\pmotif{has\_motif(C,\,late\_attn\_peak)},
    \pspan{layer\_span(C,\,$L$,\,$L'$)},
    \pcomp{comp\_ratio(C,\,attn,\,R)}, R\,>\,0.42,
    \pcomp{size(C,\,N)}, N\,<\,42.
  & 0.31 & 0.8 \\[6pt]

Location
  & \texttt{arch\_location(C)~:-}
    \newline\hspace*{6pt}\pmotif{has\_motif(C,\,late\_attn\_peak)},
    \pspan{layer\_span(C,\,$L$,\,$L'$)},
    \pcomp{comp\_ratio(C,\,attn,\,R)}, R\,>\,0.44,
    \pcomp{size(C,\,N)}, N\,<\,45.
  & 0.31 & 0.8 \\

\bottomrule
\end{tabular}

\caption{\textbf{Learned architectural signatures (Pythia-1B, 10-task pool).} Each row is a complete Horn clause; colour identifies predicate type within the conjunction. \fcolorbox{motiffg}{motifbg}{\textcolor{motiffg}{\texttt{named structural motif}}}; \fcolorbox{spanfg}{spanbg}{\textcolor{spanfg}{\texttt{normalised layer position}}}; \fcolorbox{compfg}{compbg}{\textcolor{compfg}{\texttt{circuit composition}}}. Numeric thresholds (plain text) are learned per task and not colour-coded. Confidence = F1 (one-vs-rest; 3 positive, 27 negative circuits). Complexity = $(n_\ell \times \bar{d})/10$. The \pmotif{motif} column is the only one that varies across tasks; \pspan{layer\_span} and \pcomp{composition} predicates are shared structural scaffolding present in every clause.}
\label{tab:arch-signatures-full}
\end{table*}

\begin{table*}[h]
\centering
\small
\setlength{\fboxsep}{1.2pt}
\setlength{\fboxrule}{0pt}
\setlength{\tabcolsep}{6pt}
\begin{tabular}{@{}l p{11.5cm} cc@{}}
\toprule
\textbf{Task}
  & \textbf{Learned clause $\tau_{\text{arch}}$ (Pythia-14M, 10-task pool)}
  & \textbf{Conf.} & \textbf{Cmplx.} \\
\midrule

IOI
  & \texttt{arch\_ioi(C)~:-}
    \newline\hspace*{6pt}\pmotif{has\_motif(C,\,mid\_layer\_attn\_peak)},
    \pspan{layer\_span(C,\,$L$,\,$L'$)},
    \pcomp{component\_ratio(C,\,attn,\,R)}, R\,>\,0.64,
    \pcomp{size(C,\,N)}, N\,<\,33.
  & 0.76 & 0.8 \\[6pt]

GT
  & \texttt{arch\_gt(C)~:-}
    \newline\hspace*{6pt}\pmotif{has\_motif(C,\,early\_attn\_peak)},
    \pspan{layer\_span(C,\,$L$,\,$L'$)},
    \pcomp{component\_ratio(C,\,attn,\,R)}, R\,>\,0.65,
    \pcomp{size(C,\,N)}, N\,<\,33.
  & 0.31 & 0.8 \\[8pt]

\midrule

Beneficiary
  & \texttt{arch\_beneficiary(C)~:-}
    \newline\hspace*{6pt}\pmotif{has\_motif(C,\,mid\_layer\_attn\_peak)},
    \pspan{layer\_span(C,\,$L$,\,$L'$)},
    \pcomp{component\_ratio(C,\,attn,\,R)}, R\,>\,0.61,
    \pcomp{size(C,\,N)}, N\,<\,32.
  & 0.50 & 0.8 \\[6pt]

Goal
  & \texttt{arch\_goal(C)~:-}
    \newline\hspace*{6pt}\pcomp{size(C,\,N)}, N\,<\,31.
  & 0.00 & 0.2 \\[6pt]

Instrument
  & \texttt{arch\_instrument(C)~:-}
    \newline\hspace*{6pt}\pmotif{has\_motif(C,\,mid\_layer\_attn\_peak)},
    \pspan{layer\_span(C,\,$L$,\,$L'$)},
    \pcomp{component\_ratio(C,\,attn,\,R)}, R\,>\,0.60,
    \pcomp{size(C,\,N)}, N\,<\,29.
  & 0.50 & 0.8 \\[6pt]

Location
  & \texttt{arch\_location(C)~:-}
    \newline\hspace*{6pt}\pmotif{has\_motif(C,\,early\_attn\_peak)},
    \pmotif{has\_motif(C,\,balanced\_mix)},
    \pspan{layer\_span(C,\,$L$,\,$L'$)},
    \pcomp{component\_ratio(C,\,attn,\,R)}, R\,>\,0.60,
    \pcomp{size(C,\,N)}, N\,<\,30.
  & 0.33 & 0.9 \\[6pt]

Path
  & \texttt{arch\_path(C)~:-}
    \newline\hspace*{6pt}\pmotif{has\_motif(C,\,early\_attn\_peak)},
    \pspan{layer\_span(C,\,$L$,\,$L'$)},
    \pcomp{component\_ratio(C,\,attn,\,R)}, R\,>\,0.64,
    \pcomp{size(C,\,N)}, N\,<\,31.
  & 0.31 & 0.8 \\[6pt]

Source
  & \texttt{arch\_source(C)~:-}
    \newline\hspace*{6pt}\pmotif{has\_motif(C,\,early\_attn\_peak)},
    \pspan{layer\_span(C,\,$L$,\,$L'$)},
    \pcomp{component\_ratio(C,\,attn,\,R)}, R\,>\,0.62,
    \pcomp{size(C,\,N)}, N\,<\,30.
  & 0.31 & 0.8 \\[6pt]

Time
  & \texttt{arch\_time(C)~:-}
    \newline\hspace*{6pt}\pmotif{has\_motif(C,\,balanced\_mix)},
    \pspan{layer\_span(C,\,$L$,\,$L'$)},
    \pcomp{component\_ratio(C,\,attn,\,R)}, R\,>\,0.59,
    \pcomp{size(C,\,N)}, N\,<\,30.
  & 0.31 & 0.8 \\[6pt]

Topic
  & \texttt{arch\_topic(C)~:-}
    \newline\hspace*{6pt}\pmotif{has\_motif(C,\,early\_attn\_peak)},
    \pspan{layer\_span(C,\,$L$,\,$L'$)},
    \pcomp{component\_ratio(C,\,attn,\,R)}, R\,>\,0.60,
    \pcomp{size(C,\,N)}, N\,<\,29.
  & 0.31 & 0.8 \\

\bottomrule
\end{tabular}

\caption{\textbf{Learned architectural signatures (Pythia-14M, 10-task pool).} Each row is a complete Horn clause; colour identifies predicate type within the conjunction. \fcolorbox{motiffg}{motifbg}{\textcolor{motiffg}{\texttt{named structural motif}}}; \fcolorbox{spanfg}{spanbg}{\textcolor{spanfg}{\texttt{normalised layer position}}}; \fcolorbox{compfg}{compbg}{\textcolor{compfg}{\texttt{circuit composition}}}. Numeric thresholds are learned per task and not colour-coded. Confidence = F1 on one-vs-rest. Complexity = $(n_\ell \times \bar{d})/10$. IOI is the only well-discriminated circuit; Goal degenerates to a size-only predicate.}

\label{tab:arch-signatures-14m}
\end{table*}

\begin{table*}[h]
\centering
\small
\setlength{\fboxsep}{1.2pt}
\setlength{\fboxrule}{0pt}
\setlength{\tabcolsep}{6pt}
\begin{tabular}{@{}l p{11.5cm} cc@{}}
\toprule
\textbf{Task}
  & \textbf{Learned clause $\tau_{\text{arch}}$ (LLaMA-3.2-1B, 10-task pool)}
  & \textbf{Conf.} & \textbf{Cmplx.} \\
\midrule

IOI
  & \texttt{arch\_ioi(C)~:-}
    \newline\hspace*{6pt}\pcomp{component\_ratio(C,\,attention,\,R)}, R\,>\,0.59.
  & 1.00 & 0.2 \\[6pt]

GT
  & \texttt{arch\_gt(C)~:-}
    \newline\hspace*{6pt}\pmotif{has\_motif(C,\,attn\_chain\_3)},
    \pspan{layer\_span(C,\,$L$,\,$L'$)},
    \pcomp{component\_ratio(C,\,attention,\,R)}, R\,>\,0.37,
    \pcomp{size(C,\,N)}, N\,<\,35.
  & 0.97 & 0.8 \\[8pt]

\midrule

Beneficiary
  & \texttt{arch\_beneficiary(C)~:-}
    \newline\hspace*{6pt}\pmotif{has\_motif(C,\,early\_attn\_peak)},
    \pspan{layer\_span(C,\,$L$,\,$L'$)},
    \pcomp{component\_ratio(C,\,attention,\,R)}, R\,>\,0.28,
    \pcomp{size(C,\,N)}, N\,<\,35.
  & 0.33 & 0.8 \\[6pt]

Goal
  & \texttt{arch\_goal(C)~:-}
    \newline\hspace*{6pt}\pmotif{has\_motif(C,\,scaffold\_event\_entity)},
    \pmotif{has\_motif(C,\,late\_attn\_peak)},
    \pspan{layer\_span(C,\,$L$,\,$L'$)},
    \pcomp{component\_ratio(C,\,attention,\,R)}, R\,>\,0.39,
    \pcomp{size(C,\,N)}, N\,<\,39.
  & 0.94 & 0.9 \\[6pt]

Instrument
  & \texttt{arch\_instrument(C)~:-}
    \newline\hspace*{6pt}\pmotif{has\_motif(C,\,attn\_to\_attn)},
    \pspan{layer\_span(C,\,$L$,\,$L'$)},
    \pcomp{component\_ratio(C,\,attention,\,R)}, R\,>\,0.35,
    \pcomp{size(C,\,N)}, N\,<\,38.
  & 0.94 & 0.8 \\[6pt]

Location
  & \texttt{arch\_location(C)~:-}
    \newline\hspace*{6pt}\pmotif{has\_motif(C,\,late\_attn\_peak)},
    \pspan{layer\_span(C,\,$L$,\,$L'$)},
    \pcomp{component\_ratio(C,\,attention,\,R)}, R\,>\,0.32,
    \pcomp{size(C,\,N)}, N\,<\,35.
  & 0.43 & 0.8 \\[6pt]

Path
  & \texttt{arch\_path(C)~:-}
    \newline\hspace*{6pt}\pmotif{has\_motif(C,\,mid\_layer\_attn\_peak)},
    \pmotif{has\_motif(C,\,mlp\_heavy)},
    \pspan{layer\_span(C,\,$L$,\,$L'$)},
    \pcomp{component\_ratio(C,\,attention,\,R)}, R\,>\,0.30,
    \pcomp{size(C,\,N)}, N\,<\,35.
  & 0.71 & 0.9 \\[6pt]

Source
  & \texttt{arch\_source(C)~:-}
    \newline\hspace*{6pt}\pmotif{has\_motif(C,\,mid\_layer\_attn\_peak)},
    \pspan{layer\_span(C,\,$L$,\,$L'$)},
    \pcomp{component\_ratio(C,\,attention,\,R)}, R\,>\,0.39,
    \pcomp{size(C,\,N)}, N\,<\,40.
  & 0.45 & 0.8 \\[6pt]

Time
  & \texttt{arch\_time(C)~:-}
    \newline\hspace*{6pt}\pmotif{has\_motif(C,\,late\_attn\_peak)},
    \pspan{layer\_span(C,\,$L$,\,$L'$)},
    \pcomp{component\_ratio(C,\,attention,\,R)}, R\,>\,0.28,
    \pcomp{size(C,\,N)}, N\,<\,34.
  & 0.43 & 0.8 \\[6pt]

Topic
  & \texttt{arch\_topic(C)~:-}
    \newline\hspace*{6pt}\pmotif{has\_motif(C,\,late\_attn\_peak)},
    \pspan{layer\_span(C,\,$L$,\,$L'$)},
    \pcomp{component\_ratio(C,\,attention,\,R)}, R\,>\,0.41,
    \pcomp{size(C,\,N)}, N\,<\,41.
  & 0.45 & 0.8 \\

\bottomrule
\end{tabular}

\caption{\textbf{Learned architectural signatures (LLaMA-3.2-1B, 10-task pool).} Each row is a complete Horn clause; colour identifies predicate type within the conjunction. \fcolorbox{motiffg}{motifbg}{\textcolor{motiffg}{\texttt{named structural motif}}}; \fcolorbox{spanfg}{spanbg}{\textcolor{spanfg}{\texttt{normalised layer position}}}; \fcolorbox{compfg}{compbg}{\textcolor{compfg}{\texttt{circuit composition}}}. Numeric thresholds are learned per task and not colour-coded. Confidence = F1 on one-vs-rest. Complexity = $(n_\ell \times \bar{d})/10$.}

\label{tab:arch-signatures-llama}
\end{table*}

\section{Architectural Signatures} \label{app:arch-signatures}
This appendix specifies how architectural signatures are constructed from circuit graphs. 

\subsection{Scale-Invariant Encoding}
Absolute layer indices are normalised as:
\[
    \rho(\ell) = \ell/L \in [0,1],
\]
where $\ell$ is the layer index and $L$ is the number of layers in the model. Component counts are represented as ratios rather than raw counts, for example through predicates such as \texttt{component\_ratio} and \texttt{rel\_size}. These encodings allow circuits from models with different depths and widths to be compared in a shared representational space.

The main structural predicates are:
\begin{itemize}[leftmargin=*,itemsep=1pt]
    \item \texttt{component\_ratio(C, Type, R)}: the proportion $R$ of components in circuit $C$ whose type is \texttt{Type}, e.g.\ attention heads or MLP blocks;
    \item \texttt{rel\_size(C, S)}: the size of circuit $C$ expressed relative to the model or to the circuit pool;
    \item \texttt{layer\_span(C, $L_{min}$, $L_{max}$)}: the normalised layer interval covered by circuit $C$;
    \item \texttt{has\_motif(C, M)}: whether circuit $C$ contains named motif $M$;
    \item CFS-derived predicates from Layer~2, which provide functional and causal facts about circuit components.
\end{itemize}

These predicates are designed to abstract away from model-specific identifiers while preserving structural properties relevant to mechanistic comparison. They facilitate cross-scale comparison, but are not sufficient to guarantee full mechanistic equivalence: qualitatively different computations may occur at similar relative depths in models of different scales. Transfer claims should therefore be interpreted as structural candidate retrieval followed by behavioural and causal validation, rather than as proof of mechanism identity.


\subsection{Named Motif Extraction}
Named motifs are detected from Layer~1 graph facts by deterministic topology rules over predicates such as \texttt{edge}, \texttt{type}, and \texttt{layer}. The current motif vocabulary includes:
\begin{itemize}[leftmargin=*,itemsep=1pt]
    \item \texttt{attn\_to\_attn}: a directed attention-to-attention edge;
    \item \texttt{attn\_feeds\_mlp}: an attention-to-MLP pair;
    \item \texttt{mlp\_gates\_attn}: an MLP-to-attention pair;
    \item \texttt{attn\_chain\_3}: a three-hop attention path;
    \item \texttt{attn\_mlp\_attn\_sandwich}: an attention-to-MLP-to-attention pattern;
    \item \texttt{scaffold\_event\_entity}: an attention-to-attention-to-MLP pattern associated with scaffold-mediated entity binding.
\end{itemize}
The motif vocabulary provides interpretable structural predicates for ILP$_{\text{arch}}$, while the blind structural pool described in Appendix~\ref{app:dual-pool} provides a check against over-reliance on manually named motifs.



\subsection{Dual-Pool ILP Search}
\label{app:dual-pool}
ILP$_{\text{arch}}$ searches two hypothesis pools:
\paragraph{Pool A: named motifs.} Pool~A uses the named motif vocabulary through predicates such as \texttt{has\_motif(C, M)}. These clauses are interpretable because the motif names correspond to predefined graph-topological patterns.

\paragraph{Pool B: blind structural predicates.} Pool~B enumerates raw edge-type sequences of length two and three, independently of the named motif vocabulary. A blind clause is labelled confirmatory if its edge-type pattern matches a named motif topology. Otherwise, it is flagged as a potential novel structural finding. When a novel blind clause achieves higher F1 than all named-motif clauses for a task, this indicates that the current named motif vocabulary may be incomplete for that task.

\paragraph{Clause selection.}
For each task, ILP$_{\text{arch}}$ selects the clause with the highest F1 score on the one-vs-rest classification problem. When multiple clauses obtain the same F1 score, shorter clauses are preferred. This preference implements a parsimony bias: among equally predictive hypotheses, the selected architectural signature should make the fewest structural commitments.

\subsection{Motif Scoring: Full Specification}
\label{app:motif-scoring}
Motifs are ranked before clause search using an attribution-weighted information gain score. Let $n_+$ and $n_-$ denote the number of positive and negative circuits. For motif $m$, let $\text{tp}$, $\text{fp}$, $\text{tn}$, and $\text{fn}$ denote the standard contingency counts.


\paragraph{Information gain.}
The information gain of motif $m$ is
\[
    \mathrm{IG}(m) = H(Y) - H(Y \mid m\ \text{present}),
\]
where $H$ is binary Shannon entropy using log base 2 \citep{shannon1948mathematical}. This measures how much observing motif $m$ reduces uncertainty about the task label. $\mathrm{IG}(m)$ ties at $H(Y)$ for all perfectly discriminative motifs ($\text{fp}=0$, $\text{tp}=n_+$); it does not tie in general.

\paragraph{Gain ratio.}
The gain ratio is
\[
    \mathrm{GR}(m) =
    \frac{\mathrm{IG}(m)}{H(p_{\text{with}})},
\]
where
\[
    p_{\text{with}} =
    \frac{\text{tp}+\text{fp}}{n_+ + n_-}.
\]
This corrects for rarity bias and becomes more relevant as the circuit pool grows.
\paragraph{Gini drop.} $\mathrm{GD}(m) = \mathrm{Gini}(n_+,n_++n_-) - \text{weighted Gini after split}$, where $\mathrm{Gini}(k,n)=1-(k/n)^2-(1-k/n)^2$.

This provides an alternative impurity-based measure that is less sensitive to skewed class ratios than entropy.


The final motif ranking score is
\[
    \textsc{attr\_ig}(m)
    =
    \mathrm{IG}(m) \times (1 + \bar{s}_m),
\]
where $\bar{s}_m$ is the mean attribution score of motif $m$ in the positive circuits. For two-node motifs, $\bar{s}_m$ is computed from \texttt{pair\_mean\_score}, the mean $|\text{grad}\times\text{act}|$ score over all edges of that type in positive circuits. For three-node path motifs, $\bar{s}_m$ is computed from \texttt{path3\_mean\_score}, using the mean $\sqrt{s_1s_2}$ over matching paths. Motifs absent from the edge-type attribution map receive $\bar{s}_m=0$ and therefore fall back to plain information gain.

This scoring criterion favours motifs that are both discriminative and causally salient. It does not by itself establish that a motif implements the mechanism; it only prioritises motifs for subsequent ILP clause search and validation.


\subsection{ILP Signature and Complexity Score} \label{app:ilp-signature}

The ILP signature $\mathcal{S}_{\text{ILP}}$ summarises the structural properties of the learned architectural signature $\tau_{\text{arch}}$:
\[
    \mathcal{S}_{\text{ILP}}(\tau_{\text{arch}}) =
    \langle n_{\text{clauses}},\, \bar{d},\, \mathcal{P}_H,\,
    \bar{\ell},\, \texttt{complexity} \rangle.
\]

\paragraph{$n_{\text{clauses}}$.} The number of Horn clauses in $\tau_{\text{arch}}$. In the current implementation, ILP$_{\text{arch}}$ returns a single best clause per task ($n_{\text{clauses}} = 1$); the field is retained for generality when learning clause sets.

\paragraph{$\bar{d}$: mean derivation depth.} The maximum predicate nesting depth in the clause body, measured by counting nested parentheses:
\[
    \bar{d} = \max_{\text{clause}} \max_{i} \texttt{depth}(b_i),
\]
where $\texttt{depth}(b_i)$ is the parenthesis nesting depth of literal $b_i$. For flat conjunctions of ground predicates, $\bar{d} = 1$. Deeper values arise when predicates take compound terms as arguments.

\paragraph{$\mathcal{P}_H$: predicate set.} The set of predicate names appearing in the clause body. Used as the symbolic vocabulary for the Jaccard similarity term in the signature distance metric (Appendix~\ref{app:distance-metric}). Motifs (\texttt{has\_motif} values) are stored separately as a topology discriminator, since two clauses may share predicate names while referring to structurally distinct circuit types.

\paragraph{$\bar{\ell}$: mean clause length.} The number of literals in the clause body ($\bar{\ell} = n_\ell$ for single-clause signatures), counted by parenthesis-aware splitting on top-level commas.

\paragraph{\texttt{complexity}: composite score.}
\[
    \texttt{complexity} = \frac{n_\ell \times \bar{d}}{10},
\]
where $n_\ell = \bar{\ell}$ and $\bar{d}$ is the nesting depth above. The factor of $10$ normalises to a convenient scale and has no theoretical significance. This composite penalises clauses that are simultaneously long and deeply nested; a flat 3-literal clause yields $\texttt{complexity} = 0.3$, while a 5-literal depth-2 clause yields $1.0$. \texttt{complexity} is used as a descriptive summary statistic and as a tiebreaker in the transfer fallback procedure; it is not used during clause search.

\subsection{ILP Distance Metric}
\label{app:distance-metric}
To compare two architectural signatures, we compute a distance over their ILP signatures:

\begin{equation}\small
    \begin{aligned}
    d(\mathcal{S}_1, \mathcal{S}_2) = \sqrt{
        \left(\frac{n_{\text{clauses}}^{(1)} - n_{\text{clauses}}^{(2)}}
                   {n_{\max}}\right)^2
        + \lambda \cdot \bigl(1 - J(\mathcal{P}_1, \mathcal{P}_2)\bigr)
    },
    \end{aligned}\end{equation}

where $J(\mathcal{P}_1,\mathcal{P}_2) = |\mathcal{P}_1 \cap \mathcal{P}_2| / |\mathcal{P}_1 \cup \mathcal{P}_2|$ is predicate-set Jaccard similarity and $\lambda = 0.5$.
This metric is intentionally coarse. It captures differences in clause count and predicate vocabulary, but it does not directly incorporate derivation depth $\bar{d}$, clause length $\bar{\ell}$, or the exact numeric thresholds inside predicates. As a results, two theories with identical predicate vocabularies but different proof structure may appear similar. We use this distance as an approximate structural comparison and as an acceptance criterion for fallback transfer, not as a complete measure of mechanistic equivalence.


\subsection{$\theta$-Subsumption: Operational Definition and Tractability}
\label{app:subsumption}

Clause $\gamma_1$ \emph{$\theta$-subsumes} clause $\gamma_2$, written $\gamma_1 \preceq_\theta \gamma_2$, if there exists a substitution $\theta$ mapping variables in $\gamma_1$ to terms in $\gamma_2$ such that
\[
    \gamma_1\theta \subseteq \gamma_2,
\]
treating clauses as sets of literals. Equivalently, $\gamma_1$ is at least as general as $\gamma_2$: every model of $\gamma_1$ is a model of $\gamma_2$.
The relation $\preceq_\theta$ is reflexive and transitive. When clauses are considered up to variable renaming, it supports the following mechanistic interpretations:
\begin{itemize}[leftmargin=*,itemsep=1pt]
    \item $\gamma_1 \prec_\theta \gamma_2$ (strict): $\gamma_1$ \emph{refines} $\gamma_2$ — it makes strictly stronger structural commitments, matching a proper subset of circuits;
    \item $\gamma_1 \equiv_\theta \gamma_2$ (mutual subsumption): $\gamma_1$ and $\gamma_2$ are \emph{equivalent} up to variable renaming,  they characterise the same set of circuits;
    \item $\gamma_1 \not\preceq_\theta \gamma_2$ and $\gamma_2 \not\preceq_\theta \gamma_1$: the two signatures are \emph{incomparable}, neither is a generalisation of the other.
\end{itemize}
Subsumption checking is NP-complete in general \citep{plotkin1970note,gottlob1987subsumption}. Two restrictions make it tractable in our setting.

\paragraph{Bounded clause length.} Mode declarations in ILP$_{\text{arch}}$ bound the number of literals in any learned clause body ($\bar{\ell} \leq 5$ in the current configuration). Under a fixed literal bound $k$, subsumption reduces to a matching problem of size $O(|\mathcal{P}_H|^k)$, which is polynomial for small $k$.

\paragraph{Datalog-range predicates.} All predicates in $\mathcal{B}_{\text{arch}}$ are range-restricted: every variable in a clause body appears in a positive literal. This eliminates existential variables that would otherwise exponentially expand the substitution search space.

\paragraph{Strict vs.\ Relaxed Subsumption}
Under strict $\theta$-subsumption, numeric constants must match exactly. This makes many learned clauses incomparable, even when they share the same predicate structure but differ only in thresholds. For example, clauses containing $R>0.63$ and $R>0.35$ do not match strictly because the constants differ.

For interpretability analysis, we also report \emph{relaxed} $\theta$-subsumption, where numeric constants are treated as wildcards (any threshold matches any threshold). This comparison abstracts away from exact thresholds and compares predicate structure. It should therefore be interpreted as structural generalisation, not as strict logical entailment under the original numeric constraints. 
Consider two learned clauses:
\[
\small
    \begin{aligned}
\gamma_{\text{ioi}} &: \texttt{arch\_ioi}(C) \leftarrow \texttt{component\_ratio}(C, \texttt{attn}, R),\\&\; R > 0.63. \\
\end{aligned}
\]
\[
\small
\begin{aligned}
\gamma_{\text{time}} &: \texttt{arch\_time}(C) \\&\leftarrow \texttt{has\_motif}(C, \texttt{early\_attn\_peak}),\;\\&
\texttt{layer\_span}(C, L_{\min}, L_{\max}),\\&\; \texttt{component\_ratio}(C, \texttt{attn}, R),\;\\& R > 0.35,\; \\&\texttt{size}(C, N),\; N < 36.
\end{aligned}
\]
$\gamma_{\text{ioi}} \preceq_\theta \gamma_{\text{time}}$ (relaxed): under numeric wildcarding, IOI's single literal \texttt{component\_ratio}$(C, \texttt{attn}, R)$ with $R > \_$ matches the corresponding literal in $\gamma_{\text{time}}$. Since all of $\gamma_{\text{ioi}}$'s body literals appear in $\gamma_{\text{time}}$'s body (under wildcard matching), IOI \emph{subsumes} Time. IOI is the more general signature.

$\gamma_{\text{time}} \not\preceq_\theta \gamma_{\text{ioi}}$ (relaxed): $\gamma_{\text{time}}$ contains \texttt{has\_motif}, \texttt{layer\_span}, and \texttt{size} literals that have no counterpart in $\gamma_{\text{ioi}}$. Time makes strictly stronger structural commitments.

Similarly, Time $\preceq_\theta$ Path (relaxed), because Path adds \texttt{has\_motif}$(C, \texttt{attn\_chain\_3})$ to Time's predicates. Location is \emph{incomparable} to both Time and Path because it uses different motifs (\texttt{mlp\_heavy}, \texttt{mid\_layer\_attn\_peak}).

\paragraph{Mechanistic interpretation.} 
The relaxed subsumption hierarchy reveals that attention component ratio is the minimal structural invariant shared by all circuits in Pythia-1B (captured by IOI's clause). Semantic role circuits \emph{specialise} this structural invariant by adding motif and size constraints. Within the role family, Location and Time represent \emph{distinct specialisation paths}: both refine a shared attention-ratio predicate, but through different motifs, indicating different structural realisations of binding. This is a testable claim: a circuit satisfying $\gamma_{\text{loc}}$ need not satisfy $\gamma_{\text{time}}$ (and vice versa), predicting that the location and time circuits are structurally non-interchangeable despite implementing the same operation type.
\subsection{Learned Architectural Clauses}
\label{app:learned-clauses}

Table~\ref{tab:arch-signatures-full}--\ref{tab:arch-signatures-llama} list the full set of learned clauses from the 10-task pool for all three models. Each clause is learned via one-vs-rest ILP with 3 positive circuits per task and 27 negatives. Confidence is the F1 score on this classification; complexity is $(n_\ell \times \bar{d})/10$.



\section{Formal Circuit Representation: Logical Foundation}
\label{app:owl-foundation}
The formal circuit representation (FCR) stores circuit instances as sets of ground logical facts organised across our layers. The storage substrate uses OWL~2 ontologies (providing SROIQ semantics and SPARQL-based extraction), but the logical form exposed to ILP$_{\text{arch}}$ is Prolog-style Horn clauses.

\subsection{Layers Structure}
\paragraph{Layer 0: Provenance.} Task family, evaluation corpus, annotation schema, and experimental protocol. Known from experimental design; not inferred.

\paragraph{Layer 1: Structural.} Per-node facts: \texttt{node($C$,$v$)}, \texttt{type($v$,$\tau$)},
\texttt{layer($v$,$\ell$)}. Per-edge facts: \texttt{edge($C$,$u$,$v$)}. Aggregate facts: \texttt{component\_ratio($C$,attn,$r$)}, \texttt{rel\_size($C$,$r$)}, \texttt{layer\_span($C$,$\ell_{\min}$,$\ell_{\max}$)}, \texttt{size($C$,$n$)}, \texttt{density($C$,$d$)}, \texttt{skip\_connections($C$,$k$)}, \texttt{hub\_count($C$,$k$)}, \texttt{faithfulness($C$,$f$)} where $f{=}(\mathrm{Acc}_T(C){-}\mathrm{Acc}_T^{\mathrm{abl}})/(\mathrm{Acc}_T(M){-}\mathrm{Acc}_T^{\mathrm{abl}})$~\citep{mueller2025mib}. Named motifs: \texttt{has\_motif($C$,$m$)}. Edge attribution scores are stored per edge: \texttt{edge\_score($u$,$v$,$s$)}.

\paragraph{Layer 2: Causal Functional Signature.} \texttt{dla\_score($v$,$\delta$)} and \texttt{attends\_to($v$,$r$,$p$)}, where $r$ is drawn from either the linguistic role vocabulary ($r \in \{\text{subject, object, verb, function, entity, other}\}$) or a task-specific vocabulary (e.g., $r \in \{\text{role\_filler, scaffold, other}\}$ for semantic roles). $p \in [0,1]$ is the attribution-weighted attention frequency over $\mathcal{D}_T$. Both profiles are stored; the linguistic profile is used for cross-task comparison and the task-specific profile for within-task characterisation. See Appendix~\ref{app:role-labelling} for labeller details.

\paragraph{Layer 3: Architectural Signatures.} ILP$_{\text{arch}}$ clauses of the form \texttt{arch\_$\tau$($C$) :- $b_1$, \ldots, $b_k$}, where each $b_i$ is a scale-invariant predicate from $\mathcal{B}_{\text{arch}}$.

\subsection{OWA/CWA Reconciliation}
We use OWL that operates under the Open World Assumption (OWA): absence of a fact means \emph{unknown}, not \emph{false}. ILP$_{\text{arch}}$ requires the Closed World Assumption (CWA). We reconcile this by applying a \emph{closed-world completion} at the extraction boundary: for a fixed predicate set $\mathcal{P}_{\text{target}}$ and admissible domain $\mathcal{U}_{p,i}$ per argument position, any ground atom not entailed by the OWL representation is treated as false in the ILP training database. Negation-as-failure in learned clauses is therefore interpreted relative to this completed extraction, not relative to the OWL ontology directly.

\subsection{SPARQL Extraction}
Background knowledge predicates are populated via SPARQL queries. For example, normalised depth for attention heads is computed as:
\begin{verbatim}
SELECT ?c ?depth WHERE {
  ?c rdf:type :AttentionHead .
  ?c :inLayer ?layer .
  ?c :modelDepth ?totalLayers .
  BIND(?layer/?totalLayers AS ?depth)
}
\end{verbatim}



\section{Token Role Labelling} \label{app:role-labelling}

CFS attention profiles require assigning a role label to each input token position. We use two labelling levels: a shared linguistic labeller for cross-task comparison, and task-specific labellers for within-task characterisation.

\subsection{Linguistic Roles (Shared)}
A spaCy dependency parser (\texttt{en\_core\_web\_sm}) assigns labels from the vocabulary $\{\texttt{subject}, \texttt{object}, \texttt{verb}, \texttt{function}, \texttt{entity}, \texttt{other}\}$. Tokens with dependency relations \texttt{nsubj}/\texttt{nsubjpass} are labelled \texttt{subject}; \texttt{dobj}/\texttt{pobj}/\texttt{iobj} are labelled \texttt{object}; verbs by POS tag; named entities by NER; determiners, prepositions, conjunctions, and auxiliaries are labelled \texttt{function}; all others are \texttt{other}. Subword tokens inherit the label of their aligned word via character-offset mapping. A capitalisation-based heuristic is used as fallback when spaCy is unavailable.

The \texttt{function} role separates structurally important closed-class words (determiners, prepositions) from open-class content, preventing them from being absorbed into the uninformative \texttt{other} category. A fine-grained variant (12 roles) further distinguishes \texttt{subject\_agent} from \texttt{subject\_experiencer}, \texttt{object\_patient} from \texttt{object\_theme}, \texttt{verb\_action} from \texttt{verb\_stative}, and function subtypes (\texttt{function\_prep}, \texttt{function\_det}, \texttt{function\_aux}, \texttt{function\_conj}). The coarse 6-role vocabulary is used for the cross-task results reported in the main text; the fine-grained vocabulary is available for detailed linguistic analysis.

\subsection{Task-Specific Roles}

Each task family provides a labeller that uses metadata from the contrastive pair CSV to assign task-relevant roles:

\paragraph{Semantic Role Binding.} Roles: \texttt{role\_filler} (the target word the model should predict), \texttt{scaffold\_prep} (the preposition: ``in'', ``from'', ``through''), \texttt{scaffold\_det} (the determiner within the scaffold: ``the'', ``a''), \texttt{predicate\_verb} (the main verb), \texttt{agent} (subject noun), \texttt{theme} (direct object noun), \texttt{modifier} (adjectives and adverbs), \texttt{other}. Scaffold prepositions and determiners are distinguished to reveal whether components attend to the structural cue (preposition) or its accompanying determiner. The agent and theme are identified via dependency parsing when available.

\paragraph{IOI.} Roles: \texttt{repeated\_name} (the name appearing twice, target for copying), \texttt{first\_occurrence} (its first mention), \texttt{distractor\_name} (the foil name), \texttt{io\_position} (the ``to'' before the indirect object slot), \texttt{context\_verb} (the main verb: ``gave'', ``sent''), \texttt{context\_setting} (the setting clause), \texttt{other}. Names are identified from the \texttt{target\_word} and \texttt{incorrect\_word} columns; the verb is identified via dependency parsing.

\paragraph{Greater-Than.} Roles: \texttt{first\_number} (the four-digit year, e.g., 1352), \texttt{second\_number} (the two-digit continuation, e.g., 13), \texttt{comparison\_from} (the ``from'' keyword), \texttt{comparison\_to} (the ``to'' keyword), \texttt{year\_marker} (``the year'' structural frame), \texttt{subject\_noun} (``The war''), \texttt{predicate\_verb} (``lasted''), \texttt{other}. The ``from'' and ``to'' keywords are distinguished as separate comparison markers rather than grouped into a single ``comparison\_context'' category, enabling the CFS to reveal whether components track the source, the target, or the relational structure of the comparison.

\subsection{Design Rationale}
The two-level design serves distinct purposes. The linguistic profile produces a \emph{task-agnostic} characterisation: two circuits attending primarily to \texttt{subject} tokens share a functional property regardless of whether they implement location binding or time binding. This enables the cross-task CFS clustering test (Section~\ref{sec:rq1-results}). The task-specific profile produces a \emph{task-informed} characterisation: a component attending to \texttt{scaffold} vs.\ \texttt{role\_filler} reveals whether it processes the prepositional cue or the target entity, providing finer-grained mechanistic insight within a task family.

New task families can be added by registering a labeller function mapping \texttt{(text, token\_strs, metadata)} $\to$ \texttt{List[str]} in the labeller registry, with no changes to the CFS computation or ILP pipeline.

\begin{figure*}[ht]\centering 
\includegraphics[width=\linewidth]{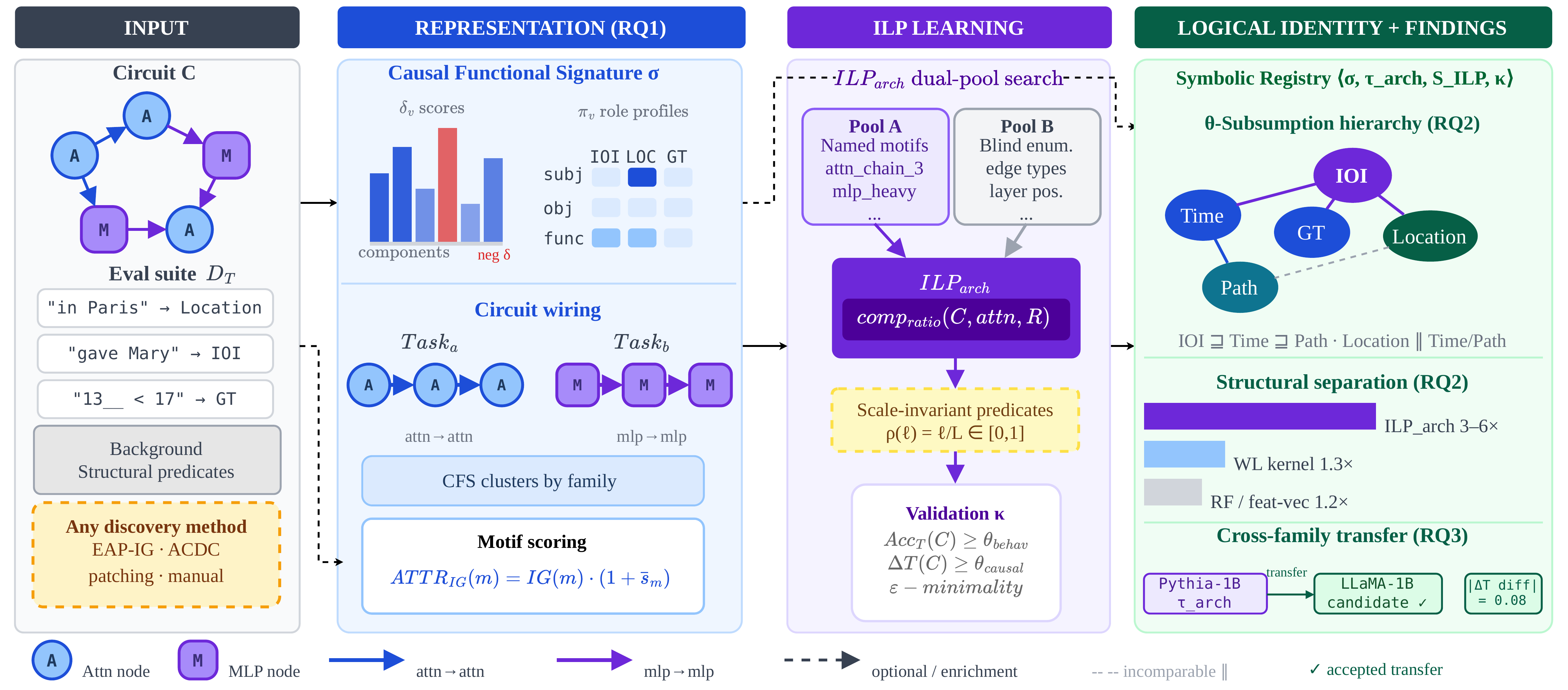} 
\caption{\textbf{Pipeline for inductive circuit theory construction.} A discovered circuit~$C$ is encoded in a Formal Circuit Representation (FCR) across four layers (provenance, structure, causal behaviour, and learned signatures). A Causal Functional Signature (CFS,~$\sigma$) is derived from causal attribution evidence (RQ1). ILP\textsubscript{arch} learns an architectural signature~$\tau_{\mathrm{arch}}$ over scale-invariant predicates. Validated triples $\langle \sigma,\, \tau_{\mathrm{arch}},\, \mathcal{S}_{\mathrm{ILP}},\, \kappa \rangle$ are stored in a Symbolic Registry supporting $\theta$-subsumption-based comparison (RQ2). Transfer Engine generates and validates candidates in a target model, with fallback to full rediscovery (RQ3). Dashed arrows indicate optional paths.} 
\label{fig:methodology_full} 
\end{figure*}

\section{Task and Model Selection}
\label{app:task-model-selection}

\subsection{Task Selection Rationale}

The three task families are chosen to cover distinct operation types while ensuring that within-family comparison is meaningful.

\paragraph{Semantic Role Binding.} Semantic roles provide a natural family of related tasks: each role (Location, Time, Path, Instrument, etc.) implements the same abstract operation, binding an entity to a thematic position within a predicate frame, but with different entities and structural cues. This enables within-family comparison: circuits implementing the same $\kappa$ should share structural properties that differ from circuits implementing a different $\kappa$. The primary evaluation uses 3 roles (Location, Time, Path) for which validated contrastive datasets exist across all three models. The scaled evaluation extends to 8 roles (adding Beneficiary, Goal, Instrument, Source, Topic) to test generalisation under increased within-family diversity.

\paragraph{IOI.} Indirect Object Identification is the most extensively studied circuit in the MI literature~\citep{wang2023interpretability,conmy2023towards}, providing a well-characterised reference mechanism with known sub-components (name movers, duplicate-token heads, S-inhibition heads). Its selection operation ($\kappa = \texttt{selection}$) is qualitatively different from binding: the circuit identifies and copies a previously mentioned entity rather than constructing a new role-filler association. This ensures the framework is tested on a genuine cross-family contrast, not merely variations of the same operation.

\paragraph{Greater-Than.} Numerical comparison introduces a non-linguistic computational domain. The comparison operation ($\kappa = \texttt{comparison}$) requires extracting numerical values and evaluating a relational predicate, distinct from both binding and selection. This tests whether the formal framework generalises beyond linguistic tasks.

\subsection{Model Cards and Selection Rationale}
Table~\ref{tab:model-cards} shows the cards for the models used in the paper. 
\paragraph{Pythia-14M and Pythia-1B.} These two models share the same architecture (GPT-NeoX), tokeniser, and training corpus (The Pile), differing only in capacity (6 vs.\ 16 layers, 4 vs.\ 8 heads). This controls for confounds: any structural differences between circuits discovered in the two models reflect differences in how the mechanism is implemented at different scales, not differences in architecture or data distribution. The 14M model also serves as a negative control for CFS: its limited capacity produces structurally homogeneous circuits (all include all MLP blocks), testing whether the framework correctly reports this uniformity rather than producing artefactual differentiation.

\paragraph{LLaMA-3.2-1B.} This model uses a different architecture (grouped-query attention, RoPE positional encoding, SiLU activation) and tokeniser (SentencePiece vs.\ GPT-NeoX BPE). It tests whether the structural patterns identified by CFS and ILP$_{\text{arch}}$, attention-dominated IOI vs.\ MLP-dominated binding, are architecture-specific properties of Pythia or general properties of how transformers implement these operations. The 1B scale matches Pythia-1B, isolating the effect of architecture from scale.

\begin{table}[h]
\centering
\small
\resizebox{\linewidth}{!}{
\begin{tabular}{@{}llll@{}}
\toprule
& \textbf{Pythia-14M} & \textbf{Pythia-1B} & \textbf{LLaMA-3.2-1B} \\
\midrule
Architecture & GPT-NeoX & GPT-NeoX & LLaMA \\
Layers & 6 & 16 & 16 \\
Heads & 4 & 8 & 32 (GQA) \\
$d_{\text{model}}$ & 128 & 2048 & 2048 \\
Parameters & 14M & 1B & 1.2B \\
Tokeniser & NeoX BPE & NeoX BPE & SentencePiece \\
Training data & The Pile & The Pile & Undisclosed \\
Positional enc. & Rotary & Rotary & RoPE \\
Activation & GELU & GELU & SiLU \\
\bottomrule
\end{tabular}
}
\caption{Model specifications}

\label{tab:model-cards}
\end{table}

\subsection{Circuit Discovery Method}

All circuits in this work are extracted using EAP-IG~\citep{hanna2024have} with $k=200$ edges and 3 integrated gradient steps. The framework is agnostic to the discovery algorithm: the formal circuit representation (Section~\ref{sec:fcr}) operates over stored nodes and attributed edges, not over the procedure that produced them. Any method that outputs a sparse subgraph with edge attribution scores, including ACDC~\citep{conmy2023towards}, path patching~\citep{goldowsky2023localizing}, activation patching~\citep{meng2022locating}, or manual circuit identification, produces valid input. We use EAP-IG because pre-validated circuits are available for our tasks across all three models; the choice does not limit the generality of the results.

\subsection{Implementation Details}\label{app:software}
Experiments were conducted using Python~3.11.13 with the following core dependencies: NumPy~1.26.4, scikit-learn~1.7.0~\cite{scikit-learn}, SciPy~1.15.3, PyTorch~2.7.1~\citep{paszke2019pytorch}, TransformerLens~2.16.1~\citep{nanda2022transformerlens}, Transformers~4.52.4, spaCy~3.8.14~\citep{honnibal2020spacy} (\texttt{en\_core\_web\_sm}), and \texttt{trace}~0.2.0 package~\citep{aljaafari-etal-2025-trace}.

\subsection{Circuit Inventory}
\label{app:circuit-inventory}
Table~\ref{tab:circuit-inventory} shows the cicuit sizes for all the tasks used in the paper. 
\begin{table}[h]
\centering
\small
{\renewcommand{\arraystretch}{0.85}
\begin{tabular}{@{}llrrrr@{}}
\toprule
\textbf{Model} & \textbf{Task} & \textbf{Nodes} & \textbf{Attn} & \textbf{MLP} & \textbf{Attn\%} \\
\midrule
\multirow{10}{*}{Py-14M} & IOI        & 26 & 20 &  6 & 77 \\
& Location    & 22 & 16 &  6 & 73 \\
& Time        & 22 & 16 &  6 & 73 \\
& Path        & 24 & 19 &  6 & 77 \\
& GT          & 27 & 21 &  6 & 78 \\
& Beneficiary & 23 & 17 &  6 & 74 \\
& Goal        & 24 & 18 &  6 & 75 \\
& Instrument  & 22 & 16 &  6 & 73 \\
& Source      & 24 & 18 &  6 & 75 \\
& Topic       & 22 & 16 &  6 & 73 \\
\midrule
\multirow{10}{*}{Py-1B} & IOI        & 38 & 26 & 11 & 70 \\
& Location    & 35 & 19 & 15 & 56 \\
& Time        & 31 & 15 & 16 & 48 \\
& Path        & 31 & 15 & 16 & 49 \\
& GT          & 41 & 25 & 16 & 62 \\
& Beneficiary & 34 & 18 & 16 & 53 \\
& Goal        & 35 & 20 & 16 & 56 \\
& Instrument  & 32 & 16 & 16 & 49 \\
& Source      & 33 & 17 & 16 & 52 \\
& Topic       & 36 & 20 & 16 & 55 \\
\midrule
\multirow{10}{*}{LL-1B} & IOI        & 42 & 30 & 12 & 71 \\
& Location    & 28 & 12 & 16 & 44 \\
& Time        & 27 & 11 & 16 & 40 \\
& Path        & 27 & 11 & 16 & 41 \\
& GT          & 29 & 14 & 15 & 48 \\
& Beneficiary & 27 & 11 & 16 & 40 \\
& Goal        & 32 & 16 & 16 & 51 \\
& Instrument  & 30 & 14 & 16 & 47 \\
& Source      & 31 & 16 & 15 & 51 \\
& Topic       & 34 & 18 & 16 & 52 \\
\bottomrule
\end{tabular}}
\caption{\textbf{Circuit inventory (30 circuits, 10 tasks $\times$ 3 models).} Values are means across 3 prompt splits. Py = Pythia, LL = LLaMA-3.2. IOI is consistently attention-dominated (70--77\%) across all models. Semantic role circuits are MLP-heavy at 1B scale (40--56\% attention); Pythia-14M circuits are uniformly attention-heavy (73--78\%) due to the small model's limited MLP capacity.}
\label{tab:circuit-inventory}
\end{table}


\section{Linguistic Grounding} \label{app:linguistic-grounding}

The operation types that emerge from CFS and ILP$_{\text{arch}}$ ($\kappa = \texttt{binding}$ vs.\ $\kappa = \texttt{selection}$) correspond to well-studied distinctions in formal linguistics.

\paragraph{Binding and $\theta$-role assignment.}
In formal semantics, $\theta$-role assignment is the process by which a predicate assigns thematic roles (Agent, Patient, Location, Time, etc.) to its arguments~\citep{chomsky1981lectures}. The semantic role circuits we study implement a computational analogue: the circuit \emph{binds} an entity (the role filler) to a structural position within the predicate frame, mediated by syntactic markers (prepositions and determiners). CFS confirms this interpretation: at the 1B scale, binding circuits attend primarily to \texttt{subject} positions (the entity being bound) and to \texttt{scaffold} positions (the structural cue that signals which role is being assigned). The MLP-dominated edge composition (64--96 mlp$\to$mlp edges) suggests that binding is implemented through feature transformation rather than token copying, consistent with the view that $\theta$-role assignment is a compositional operation that constructs new representations rather than merely moving existing ones.

\paragraph{Selection and referential processing.}
IOI circuits implement a qualitatively different operation: identifying a previously mentioned entity (the indirect object) and copying it to the prediction position. This is closer to anaphora resolution or coreference than to argument structure. CFS confirms the distinction: IOI components attend to positional rather than thematic features (71--76\% \texttt{other}-dominant), and the circuit is dominated by attention$\to$attention edges (85 in Pythia-1B), reflecting a copy-via-attention mechanism. The task-specific profile reveals that 10 of 27 causal components attend to \texttt{repeated\_name} positions, directly implementing the ``find and copy the repeated token'' strategy identified in prior work~\citep{wang2023interpretability}.

\paragraph{Greater-Than and relational comparison.}
Greater-Than circuits occupy an intermediate structural position: they share the predicate vocabulary of semantic role circuits (\texttt{has\_motif}, \texttt{layer\_span}, \texttt{size}) but use a more balanced attention/MLP composition (59\% attention in Pythia-1B). This is consistent with numerical comparison requiring both relational processing (comparing two quantities, analogous to predicate-argument structure) and feature extraction (identifying the numerical tokens, analogous to entity recognition). The task-specific profile shows attention to \texttt{comparison\_context} tokens (``from'', ``to''), suggesting that the circuit tracks the relational structure of the comparison rather than simply extracting digit values.

\paragraph{Implications.}
The alignment between our computationally derived categories (binding, selection, comparison) and linguistically motivated distinctions ($\theta$-role assignment, referential processing, relational semantics) is not a design choice; it emerges from the data. CFS and ILP$_{\text{arch}}$ were not given linguistic priors; the operation types were assigned post hoc based on the task definitions. That the resulting structural patterns (MLP-mediated vs.\ attention-mediated vs.\ hybrid) mirror the functional distinctions posited by linguistic theory provides convergent evidence. This evidence indicates that models appear to have developed computational strategies that reflect genuine properties of language processing, not merely surface statistical regularities.

\section{Hyperparameters and Thresholds}
\label{app:hyperparameters}\label{app:implementation}
Table~\ref{tab:hyperparameters} lists the threshold parameters used in the validation criteria (Section~\ref{sec:validation}) and transfer procedure (Section~\ref{sec:transfer}).

\begin{table}[h]
\centering
\small
\resizebox{\linewidth}{!}{
\begin{tabular}{llp{6cm}}
\toprule
\textbf{Parameter} & \textbf{Default} & \textbf{Rationale} \\
\midrule
$\theta_{\text{behav}}$ & $0.50$ & Minimum fraction of $\mathcal{D}_T$ on which the circuit supports the target behaviour. Set conservatively to accommodate circuits in small models where faithfulness may be limited. \\
$\theta_{\text{causal}}$ & $0.10$ & Minimum task-performance drop upon circuit ablation. Set low to retain circuits whose causal contribution is positive but small. \\
$\epsilon$ & $0.20$ & Approximate minimality tolerance for greedy edge ablation. A circuit is treated as approximately minimal if no greedy removal path can reduce the circuit by more than an $\epsilon$ fraction while preserving both behavioural and causal validation criteria. \\
$\epsilon_{\text{DLA}}$ & $0.05$ & DLA score below which a component is flagged as structurally present but causally marginal in the CFS. \\
$\delta_\rho$ & $1/L_\beta$ & Depth tolerance for candidate generation during transfer: one layer in the target model. This tolerance scales automatically with target-model depth. \\
$\delta_{\text{ILP}}$ & $0.30$ & Maximum ILP signature distance (Appendix~\ref{app:distance-metric}) for the transfer fallback procedure. Sensitivity analysis on the 10-task pool (Appendix~\ref{app:sensitivity}) finds optimal F1 at $\delta_{\text{ILP}}=0.10$; the default $0.30$ is a conservative upper bound. \\
$\lambda$ & $0.50$ & Weight on the predicate-set Jaccard term in the ILP signature distance metric (Appendix~\ref{app:distance-metric}). \\
\bottomrule
\end{tabular}
}
\caption{Hyperparameters and threshold definitions. All values are defaults used in our experiments; they may require adjustment for tasks with very small evaluation suites, different ablation protocols, or circuits of atypical size.}
\label{tab:hyperparameters}
\end{table}
\paragraph{Threshold selection.}
The validation thresholds $\theta_{\text{behav}}$ and $\theta_{\text{causal}}$ are not optimised on held-out data. They are set conservatively to retain circuits with positive but small causal contributions. We perform a limited sensitivity analysis for the transfer distance threshold $\delta_{\text{ILP}}$ in Appendix~\ref{app:sensitivity}; broader sensitivity analysis over all validation thresholds is left as a limitation.
\paragraph{Random-subgraph significance tests.}
For each validated circuit, we compare the observed causal effect $\Delta_T(C)$ against size-matched random subgraphs sampled from the same model. Statistical significance is assessed using permutation tests at $\alpha=0.05$. This check ensures that the observed causal degradation is not explained merely by ablating a subgraph of comparable size. In the evaluated settings, random circuits yield $\Delta_T \approx 0$, supporting the interpretation that validated circuits capture task-relevant causal structure rather than generic ablation sensitivity.

\section{Additional Results}\label{app:additional-results}
\subsection{Full ILP Distance Matrix (10 Tasks)}
\label{app:full-distance}
\begin{figure*}[h]
\centering
\includegraphics[width=0.8\linewidth]{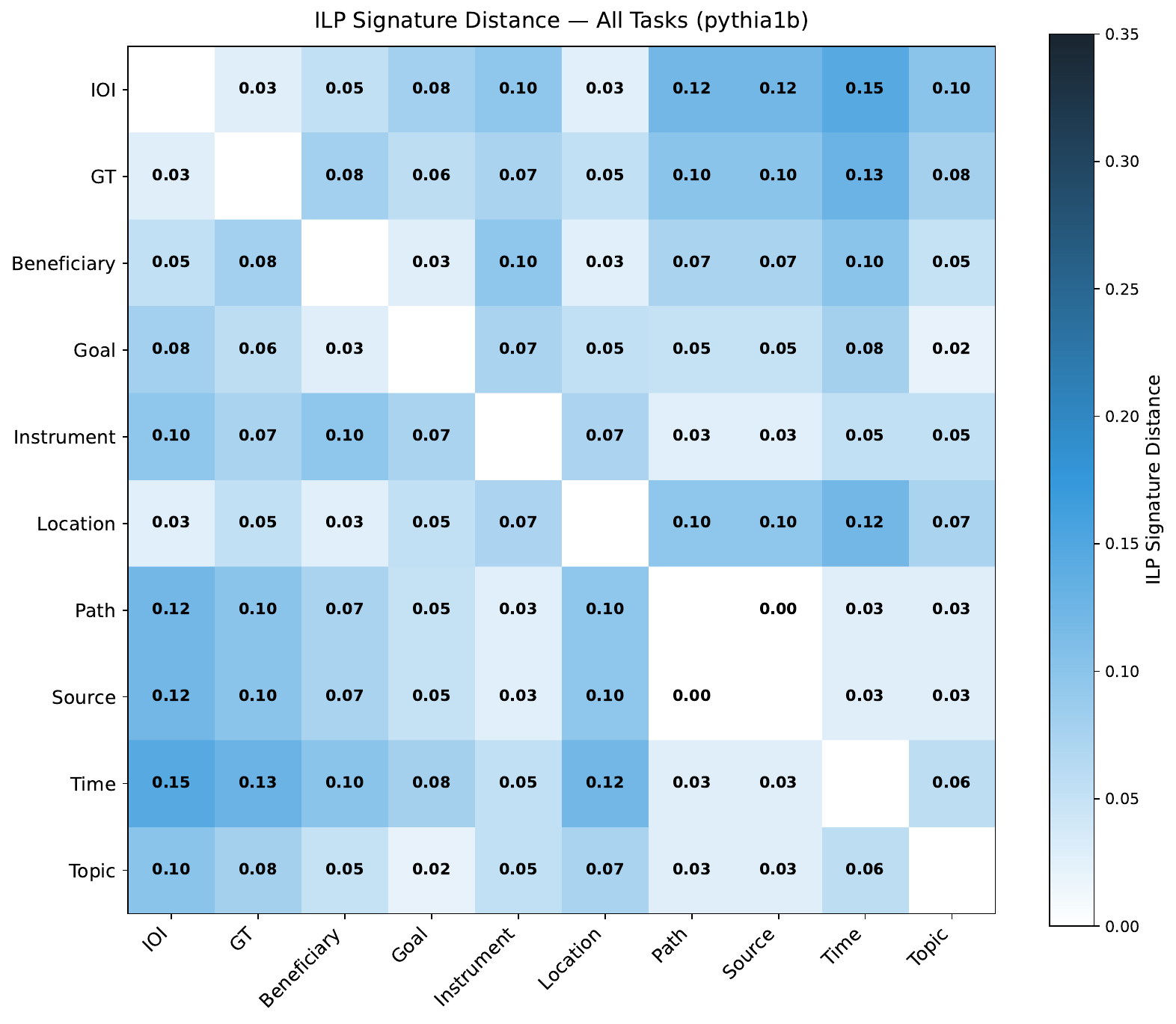}
\caption{\textbf{Full ILP signature distance (Pythia-1B, 10 tasks).} IOI and GT retain high separation from the role cluster. Within roles, Source and Time form a tight sub-cluster, while Goal is the most distant role. The expanded matrix confirms that the 5-task results (Table~\ref{tab:distance-matrix}) generalise: IOI remains structurally distinct across all task comparisons.}
\label{fig:app-distance-full}
\end{figure*}

\begin{figure*}[h]
\centering
\includegraphics[width=0.8\linewidth]{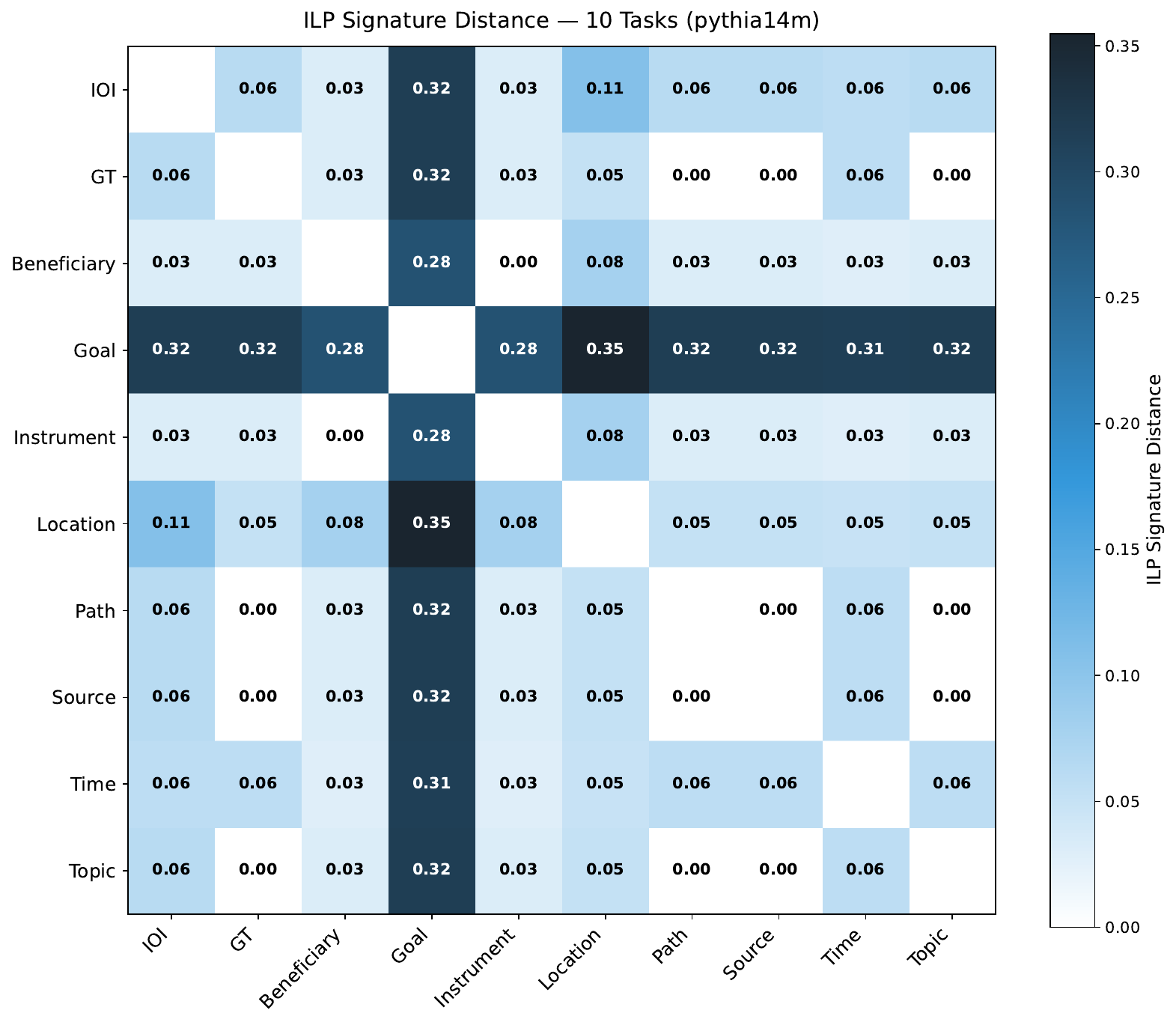}
\caption{\textbf{ILP signature distance (Pythia-14M, 10 tasks).} Structural homogeneity dominates: most pairs cluster at $d \leq 0.06$, with many identical ($d = 0.00$). Goal is the sole outlier ($d = 0.28$--$0.36$), driven by its degenerate size-only clause (Table~\ref{tab:arch-signatures-14m}). No pair outside Goal clears $\delta_{\text{ILP}} = 0.30$.}
\label{fig:distance-14m}
\end{figure*}

\begin{figure*}[h]
\centering
\includegraphics[width=0.8\linewidth]{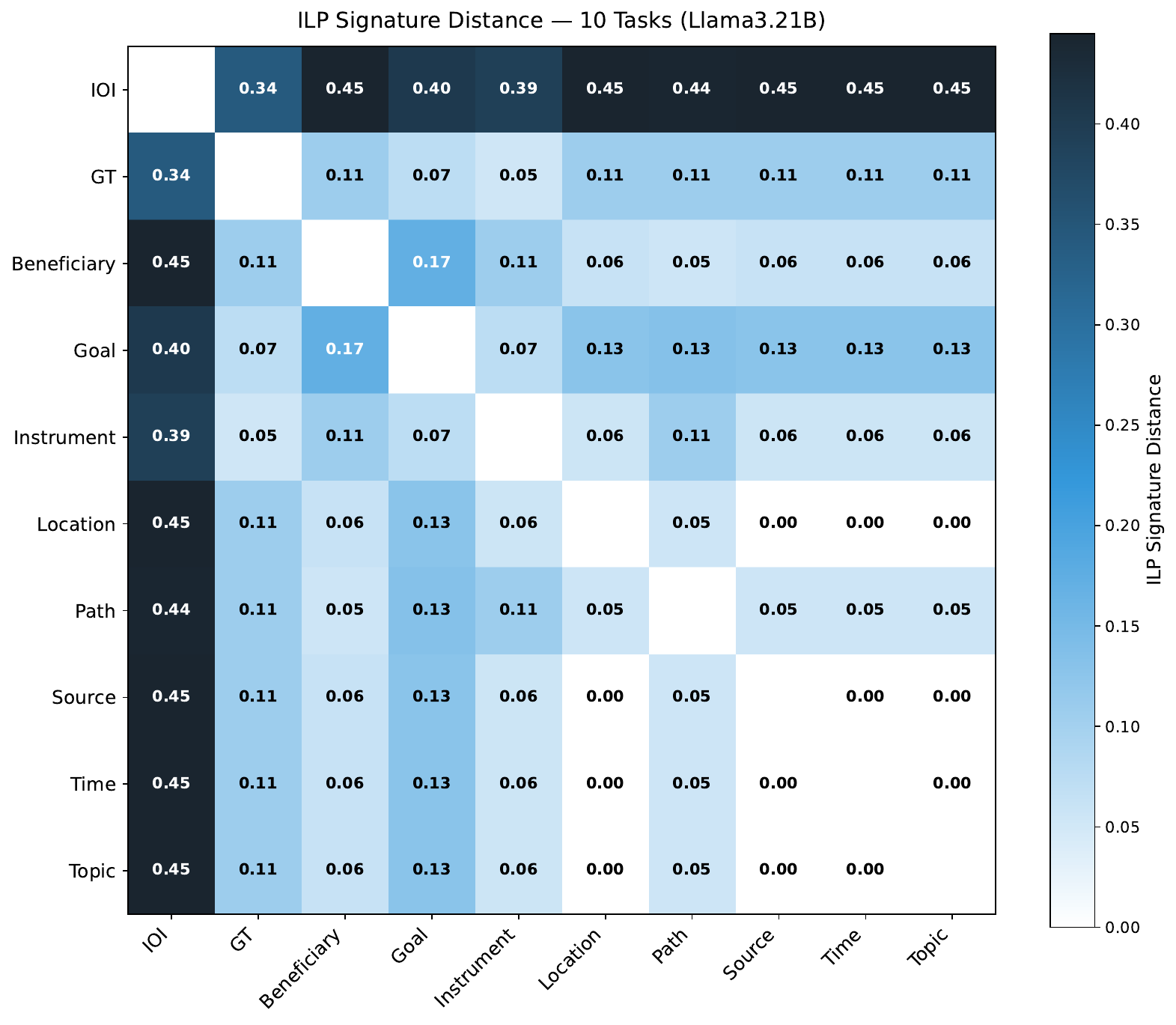}
\caption{\textbf{ILP signature distance (LLaMA-3.2-1B, 10 tasks).} IOI is strongly separated from all others ($d = 0.34$--$0.45$). GT is intermediate ($d = 0.05$--$0.11$ from roles). Goal is distinctive ($d = 0.07$--$0.17$) owing to its unique \texttt{scaffold\_event\_entity} motif. Location, Time, Source, and Topic share identical signatures ($d = 0.00$), collapsing into one structural class.}
\label{fig:distance-llama}
\end{figure*}

Figures~\ref{fig:app-distance-full}, \ref{fig:distance-14m}, and~\ref{fig:distance-llama} show the full 10$\times$10 ILP signature distance matrices for all three models. The distance metric uses clause count, predicate-set Jaccard, and motif-set Jaccard (Appendix~\ref{app:distance-metric}).

\begin{table}[t]
\centering
\small
\begin{tabular}{@{}lccccc@{}}
\toprule
& \textbf{IOI} & \textbf{LOC} & \textbf{PATH} & \textbf{TIME} & \textbf{GT} \\
\midrule
\textbf{IOI}  & ,  & 0.35 & 0.38 & 0.45 & 0.32 \\
\textbf{LOC}  & 0.35 & ,  & 0.08 & 0.13 & 0.05 \\
\textbf{PATH} & 0.38 & 0.08 & ,  & 0.11 & 0.08 \\
\textbf{TIME} & 0.45 & 0.13 & 0.11 & ,  & 0.13 \\
\textbf{GT}   & 0.32 & 0.05 & 0.08 & 0.13 & ,  \\
\bottomrule
\end{tabular}
\caption{\textbf{ILP signature distance (Pythia-1B).} Two clusters emerge: IOI is distant from all other circuits ($d = 0.32$--$0.45$), while semantic roles and Greater-Than form a tight cluster ($d = 0.05$--$0.13$). GT's structural proximity to role circuits reflects its learned clause using the same predicate set (\texttt{has\_motif}, \texttt{layer\_span}, \texttt{size}, \texttt{component\_ratio}) despite implementing a different computation. The transfer threshold $\delta_{\text{ILP}} = 0.30$ cleanly separates IOI from all other circuits.}
\label{tab:distance-matrix}
\end{table}

\subsubsection{Within-Task Structural Variance}
\label{app:within-task-variance}

To assess whether circuits discovered from different prompt subsets for the \emph{same task} are structurally consistent, we compute pairwise WL kernel distances among the 3 splits per task (within-task) and compare to the mean distance to other tasks' circuits (between-task). The ratio (between/within) measures how much more similar same-task circuits are to each other than to different-task circuits. On LLaMA-3.2-1B, the IOI within/between ratio is 2.3$\times$, the highest across all models and tasks, confirming that the IOI attention-chain topology is a robust structural invariant of the architecture, not an artefact of the specific prompts used for discovery. For Pythia-14M, all tasks have ratios $\approx 1.0$, confirming the model's structural homogeneity.

\begin{figure*}[h]
\centering
\includegraphics[width=0.75\linewidth]{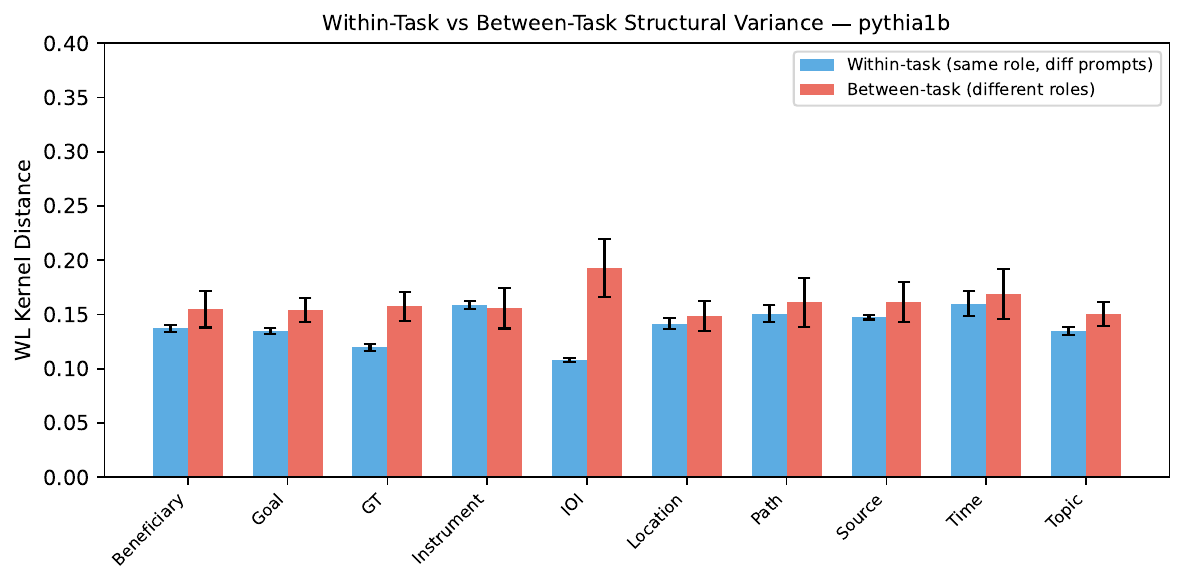}
\caption{\textbf{Within-task vs.\ between-task structural variance (Pythia-1B).} Each bar pair shows the mean WL kernel distance among splits of the same task (blue, within) vs.\ to other tasks (red, between). IOI has the largest gap (1.8$\times$), confirming its prompt-invariant circuit topology. Most semantic roles show ratios near 1.0, indicating that prompt subsets produce circuits as structurally different from each other as from other tasks, consistent with the lower ILP confidence for these roles.}
\label{fig:app-variance}
\end{figure*}
\begin{figure}[t!]
\centering
\includegraphics[width=0.8\linewidth]{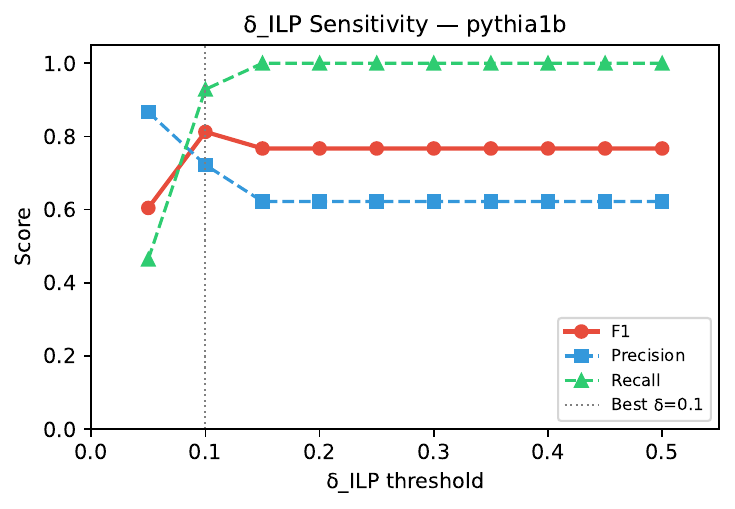}
\caption{\textbf{$\delta_{\text{ILP}}$ sensitivity (Pythia-1B, 10 tasks).} F1 peaks at $\delta_{\text{ILP}} = 0.10$ (F1 = 0.81) with precision 0.72 and recall 0.93. The curve is stable across 0.10--0.15; beyond 0.15, all same-family pairs are accepted (recall = 1.0) but precision drops as cross-family pairs also fall below threshold.}
\label{fig:app-sensitivity}
\end{figure}

\subsection{$\delta_{\text{ILP}}$ Sensitivity Analysis}
\label{app:sensitivity}

Figure~\ref{fig:app-sensitivity} shows the transfer-pair classification performance as the acceptance threshold $\delta_{\text{ILP}}$ is varied from 0.05 to 0.50. A pair is ``same-family'' if both circuits share a task family label and ``cross-family'' otherwise. Accepting a pair (predicting same-family) when $d < \delta_{\text{ILP}}$ yields the precision/recall/F1 curve.


\subsection{Edge Type Distribution}
\label{app:edge-types}
Table~\ref{tab:edge-types-full} breaks down the 200 retained edges per circuit by type (attn$\to$attn, attn$\to$mlp, mlp$\to$attn, mlp$\to$mlp) for both 1B-scale models. The distribution quantifies the wiring distinction identified by CFS in Section~\ref{sec:rq1-results}.

\begin{table}[h]
\centering
\small
\begin{tabular}{@{}lrrrrr@{}}
\toprule
\textbf{Task} & \textbf{a$\to$a} & \textbf{a$\to$m} & \textbf{m$\to$a} & \textbf{m$\to$m} & \textbf{Attn\%} \\
\midrule
\multicolumn{6}{@{}l@{}}{\emph{Pythia-1B}} \\
IOI        & 85 &  19 & 44 & 11 & 73\% \\
Location   &  4 &  37 & 39 & 67 & 49\% \\
Time       &  1 &  19 & 27 & 96 & 45\% \\
Path       &  7 &  24 & 53 & 64 & 54\% \\
GT         & 25 &  34 & 47 & 44 & 59\% \\
\midrule
\multicolumn{6}{@{}l@{}}{\emph{LLaMA-3.2-1B}} \\
IOI        & 44 &  15 & 33 & 10 & 69\% \\
Location   &  2 &  18 & 23 & 85 & 47\% \\
Time       &  0 &   9 & 16 &105 & 37\% \\
Path       &  0 &  11 & 19 & 80 & 41\% \\
GT         &  9 &  14 & 27 & 59 & 47\% \\
\bottomrule
\end{tabular}
\caption{\textbf{Edge type distribution (1B-scale models).} IOI has 85/44 attn$\to$attn edges in Pythia/LLaMA; semantic role circuits have 64--105 mlp$\to$mlp edges. LLaMA Time has zero attn$\to$attn edges.}
\label{tab:edge-types-full}
\end{table}

\subsection{Baseline Comparison Details}
\label{app:baseline-details}

ILP signature distance achieves ${\sim}3\times$ better IOI separation than the WL kernel on Pythia-1B and $2.7\times$ better on LLaMA-1B ($4.15\times$ vs.\ $1.53\times$). The WL kernel correlates with ILP ($r = 0.81$--$0.95$) but captures only local graph topology, not conjunctive predicate structure. The random forest has access to the same features as ILP, but does not produce compact inspectable clauses over them; its 60\% LOO accuracy (misclassifying GT and IOI) indicates that explicit logical structure contributes useful discrimination beyond the feature inventory alone. On Pythia-14M, no method achieves meaningful separation (all $\leq1.1\times$), confirming structural homogeneity as a genuine property.







\subsection{Random Circuit Baseline}
\label{app:random-baseline}
For each circuit in the corpus, 50 size-matched random subgraphs are sampled: each random subgraph contains the same number of nodes $|V|$ and edges $|E|$ as the real circuit, selected uniformly at random from the model graph. Each random subgraph is evaluated using the same CNP metric as the real circuit, producing a causal contribution score $\Delta_T^{\text{rand}} = \text{AccT}(M^{\text{rand}}) - \text{AccT}(M^{\text{ablate}})$, where $M^{\text{ablate}}$ denotes the model with all circuit edges zeroed. A one-sided permutation test then checks whether the real circuit's $\Delta_T$ exceeds the empirical distribution of $\Delta_T^{\text{rand}}$ at $p < 0.05$. Table~\ref{tab:random-baseline-summary} reports results across 34 circuits for Pythia-14M, 35 for Pythia-1B, and 33 for LLaMA-3.2-1B. Random subgraphs consistently yield $\Delta_T^{\text{rand}} \approx 0$, confirming that causal contribution is a property of the discovered circuit structure rather than of circuit size.

\begin{table}[h]
\centering
\small
\begin{tabular}{@{}lrrrr@{}}
\toprule
\textbf{Model} & \textbf{Circuits} & \textbf{Sig.} & \textbf{Non-sig.} & \textbf{Rate} \\
\midrule
Pythia-14M   & 34 & 29 & 5 & 85\% \\
Pythia-1B    & 35 & 28 & 7 & 80\% \\
LLaMA-3.2-1B & 33 & 26 & 7 & 79\% \\
\bottomrule
\end{tabular}
\caption{\textbf{Random circuit baseline summary.} For each model, the corpus of discovered circuits is tested against 50 size-matched random subgraphs per circuit. Sig.\ = real circuit $\Delta_T$ significantly exceeds the random distribution ($p < 0.05$, one-sided permutation test).}
\label{tab:random-baseline-summary}
\end{table}
\begin{table*}[h]
\centering
\small
\begin{tabular}{@{}ll|cclrr|cclrr@{}}
\toprule
& & \multicolumn{5}{c|}{\textbf{Pythia-14M $\to$ Pythia-1B}} & \multicolumn{5}{c}{\textbf{LLaMA-3.2-1B $\to$ Pythia-1B}} \\
\textbf{Task} & $\kappa$ & \textbf{Acc.} & \textbf{N} & \textbf{Best Match} & $\Delta_T^{\text{src}}$ & $\Delta_T^{\text{tgt}}$ & \textbf{Acc.} & \textbf{N} & \textbf{Best Match} & $\Delta_T^{\text{src}}$ & $\Delta_T^{\text{tgt}}$ \\
\midrule
IOI & sel. & 8/41 & \checkmark & IOI & $-$0.91 & 0.65 & 8/41 & \checkmark & IOI & 5.24 & 0.65 \\
GT & comp. & 2/37 & \checkmark & GT & 7.95 & 3.39 & 2/41 & \checkmark & GT & 1.23 & 3.39 \\
Location & bind. & 22/29 & \checkmark & Goal & 1.93 & 1.00 & 25/41 & \checkmark & Goal & 1.20 & 1.00 \\
Path & bind. & 19/36 & \checkmark & Goal & 0.53 & 0.81 & 19/41 & \checkmark & Goal & 0.48 & 0.81 \\
Time & bind. & 3/32 & \checkmark & Goal & 0.75 & 0.12 & 3/41 & \checkmark & Goal & $-$0.85 & 0.12 \\
\bottomrule
\end{tabular}
\caption{\textbf{Live transfer results.} Acc.\ = accepted/total candidates; N = task-correct match (\checkmark); $\Delta_T^{\text{src}}$ = source circuit's causal contribution; $\Delta_T^{\text{tgt}}$ = best accepted candidate's causal contribution on target data. $\kappa$: sel.\ = selection, comp.\ = comparison, bind.\ = binding. \S~The target registry holds three prompt-split circuits per task; the best accepted IOI circuit captures a portion of the full Py-1B IOI computation ($\Delta_T = 4.37$, Table~\ref{tab:cross-scale-dt}).}
\label{tab:live-transfer-full}
\end{table*}
\paragraph{Non-significant cases.} 
All non-significant circuits have negative or marginal $\Delta_T$ values, meaning the circuit's pathway performs at or below the ablated baseline; this reflects genuine properties of those circuits rather than a failure of the methodology. On Pythia-14M (5 non-significant): the three IOI circuits (primary and splits s0, s1) have $\Delta_T \in \{-0.72, -0.81, -0.60\}$, consistent with the known sign reversal of the IOI causal contribution at the 14M scale. Two Beneficiary splits (s0: $-0.21$; s1: $+0.13$, marginally non-significant) show near-zero contributions. On {Pythia-1B} (7 non-significant): three Time circuits (primary: $-0.02$; s0: $-0.31$; s2: $-0.13$) and one Path split (s1: $-0.02$) have negative $\Delta_T$, and three Beneficiary splits (s0: $-0.11$; s1: $-0.04$; s2: $-0.21$) contribute below the ablated baseline. On {LLaMA-3.2-1B} (7 non-significant): all four Time circuits (primary: $-0.75$; s0: $-0.51$; s1: $-0.34$; s2: $-0.75$) and three Beneficiary splits (s0: $-0.14$; s1: $-0.03$; s2: $-0.01$) are non-significant, following the same Time sign-reversal and Beneficiary marginal-contribution patterns seen across all three models.

\subsection{Live Transfer Results}
\label{app:live-transfer}

We evaluate end-to-end transfer for two source$\to$target pairs, both targeting Pythia-1B as the recipient: (1)~Pythia-14M$\to$Pythia-1B (cross-scale, same family, $70\times$ parameter gap) and (2)~LLaMA-3.2-1B$\to$Pythia-1B (cross-family, different architecture and tokeniser). The transfer engine searches the Pythia-1B registry (10 tasks $\times$ 3 splits = 30 circuits) for candidates matching the source circuit's $\tau_{\text{arch}}$, then applies behavioural ($\theta_{\text{behav}} = 0.50$) and causal ($\theta_{\text{causal}} = 0.10$) validation. Table~\ref{tab:live-transfer-full} shows that all 5 tasks are accepted for both transfer directions. Three patterns emerge:

\paragraph{Task-correct transfer for IOI and GT.} Both IOI and GT find their exact task match in the target registry, despite having very different source $\Delta_T$ values ($-0.91$ vs.\ $5.24$ for IOI across the two sources). The selectivity is high: only 2/37--41 candidates pass for GT (both are GT circuits from different prompt splits), and 8/41 for IOI. GT candidates that fail are rejected at the behavioural threshold ($\theta_{\text{behav}} = 0.50$), confirming that the structural signature is discriminative.

\paragraph{Family-correct transfer for binding tasks.} All three semantic role tasks (Location, Path, Time) select Goal as their best candidate, a role not present in the 5-task source pool. This is semantically valid: Goal implements binding via the same motif vocabulary (\texttt{mlp\_heavy}, \texttt{early\_attn\_peak}), and its $\Delta_T = 1.00$ is the highest among binding circuits in Pythia-1B. The acceptance rate varies by task: Location accepts 22--25 candidates (structurally similar to many roles), while Time accepts only 3 (its clause is more restrictive). Rejected candidates fail primarily at the causal threshold, not the behavioural one.

\paragraph{Identical outcomes across transfer directions.} The best accepted candidate and its $\Delta_T$ are identical for both source models, because the target registry is the same (Pythia-1B). What differs is the source $\Delta_T$: IOI's sign reversal ($-0.91$ in 14M vs.\ $5.24$ in LLaMA) and Time's negative source value in LLaMA ($-0.85$) confirm that the transfer engine correctly identifies structurally matched candidates regardless of source behaviour. The accept/reject pattern is architecture-independent.

\section{Formal Algorithms}
\label{app:algorithms}
Algorithm~\ref{alg:characterisation} formalises the circuit characterisation pipeline described in Section~\ref{sec:methodology}: given a discovered circuit $C$ and a populated registry $\mathcal{R}$, it derives the logical identity \(\mathcal{L}(C)=\langle\sigma,\,\tau_{\mathrm{arch}},\,\mathcal{S}_{\mathrm{ILP}},\,\kappa\rangle\) across four layers (provenance, structural predicates, causal functional signature, and ILP-learned architectural signature). Algorithm~\ref{alg:transfer} formalises the transfer procedure described in Section~\ref{sec:methodology}: given a source logical identity and a target model $M_\beta$, it retrieves structurally compatible candidates from the registry, applies behavioural and causal screening, and falls back to full rediscovery if no candidate passes.  \begin{algorithm}[t]
  \small
  \caption{Circuit Characterisation Pipeline}
  \label{alg:characterisation}
  \begin{algorithmic}[1]
  \Require Circuit $C=(V_C,E_C)$, evaluation suite $\mathcal{D}_T$,
           circuit registry $\mathcal{R}$, background knowledge $\mathcal{B}$
  \Ensure  Logical identity $\mathcal{L}(C)=\langle\sigma,\tau_{\text{arch}},\mathcal{S}_{\text{ILP}},\kappa\rangle$

  \Statex \textbf{Layer 0 — Provenance}
  \State Record $\langle\texttt{task},\texttt{model},\texttt{discovery\_method}\rangle$ as ground facts $\text{prov}(C)$

  \Statex \textbf{Layer 1 — Structure and named motifs}
  \For{each component $v\in V_C$ at layer $\ell$}
      \State Assert $\texttt{node}(C,v)$, $\texttt{type}(v,\tau)$, $\texttt{layer}(v,\ell)$
  \EndFor
  \State Compute and assert $\texttt{layer\_span}(C,\ell_{\min},\ell_{\max})$,
         $\texttt{component\_ratio}(C,\tau,r)$, $\texttt{rel\_size}(C,s)$
  \State Extract named motifs $\mathcal{M}(C)\subseteq\{\texttt{attn\_chain\_3},\texttt{mlp\_heavy},\ldots\}$;
         assert $\texttt{has\_motif}(C,m)$ for each $m\in\mathcal{M}(C)$
  \State Store Layer~1 facts in $\mathcal{B}$

  \Statex \textbf{Layer 2 — Causal Functional Signature (CFS)}
  \For{each node $v\in V_C$}
      \State $\delta_v\gets\text{DLA}(v,\mathcal{D}_T)$
      \If{$|\delta_v|<\epsilon_{\text{DLA}}$}
          \State Flag $v$ as causally marginal
      \Else
          \State Compute attribution-weighted attention profile $\pi_v$
          \State Label positions with linguistic roles $\pi_v^{\text{ling}}$ and task roles $\pi_v^{\text{task}}$
      \EndIf
  \EndFor
  \State $\sigma\gets\{\langle v,\delta_v,\pi_v^{\text{ling}},\pi_v^{\text{task}}\rangle:v\in V_C,\;|\delta_v|\ge\epsilon_{\text{DLA}}\}$
  \State Store Layer~2 CFS facts in $\mathcal{B}$

  \Statex \textbf{Layer 3 — ILP\textsubscript{arch}: learning $\tau_{\text{arch}}$}
  \State $E^+\gets\{C'\in\mathcal{R}:\text{prov}(C').\texttt{task}=\text{prov}(C).\texttt{task}\}$
  \State $E^-\gets\{C'\in\mathcal{R}:\text{prov}(C').\texttt{task}\neq\text{prov}(C).\texttt{task}\}$
  \State Rank named motifs: $\textsc{attr\_ig}(m)\gets\mathrm{IG}(m)\times(1+\bar{s}_m)$
  \State $\text{Pool}_A\gets$ hypothesis space over \texttt{has\_motif} predicates
         \Comment{named motifs}
  \State $\text{Pool}_B\gets$ hypothesis space over \texttt{edge}, \texttt{type}, \texttt{layer},
  \Statex \hspace{4.5em} \texttt{component\_ratio}, \texttt{layer\_span}, \texttt{skip\_connections}, \texttt{hub\_count}
  \Statex \hspace{4.5em} \Comment{blind structural predicates; no \texttt{has\_motif}}
  \State $\mathcal{H}_A\gets\textsc{ILP}(E^+,E^-,\mathcal{B},\text{Pool}_A)$
  \State $\mathcal{H}_B\gets\textsc{ILP}(E^+,E^-,\mathcal{B},\text{Pool}_B)$
  \State $\tau_{\text{arch}}\gets\arg\max_{\gamma\,\in\,\mathcal{H}_A\cup\mathcal{H}_B}\mathrm{F1}(\gamma)$,
         breaking ties by $|\gamma|$
  \State $\mathcal{S}_{\text{ILP}}\gets\langle n_{\text{clauses}},\bar{d},\mathcal{P}_H,\bar{\ell},\texttt{complexity}\rangle$

  \Statex \textbf{Validation}
  \State Compute $\text{Acc}_T(C)$ and $\Delta_T(C)$ on $\mathcal{D}_T$
  \State $\kappa\gets\langle\text{Acc}_T(C),\Delta_T(C),\text{is\_minimal}(C)\rangle$
  \State Register $\mathcal{L}(C)=\langle\sigma,\tau_{\text{arch}},\mathcal{S}_{\text{ILP}},\kappa\rangle$ in $\mathcal{R}$
  \State \Return $\mathcal{L}(C)$
  \end{algorithmic}
  \end{algorithm}

  \begin{algorithm}[t]
  \small
  \caption{Mechanistic Knowledge Transfer}
  \label{alg:transfer}
  \begin{algorithmic}[1]
  \Require Source identity
           $\mathcal{L}(C_\alpha)=\langle\sigma_\alpha,
           \tau_{\mathrm{arch},\alpha},\mathcal{S}_\alpha,\kappa_\alpha\rangle$;
           target model $M_\beta$ with depth $L_\beta$;
           registry $\mathcal{R}$;
           evaluation suite $\mathcal{D}_T^\beta$;
           thresholds $\theta_{\mathrm{behav}},\theta_{\mathrm{causal}},\delta_{\mathrm{ILP}}$
  \Ensure  Accepted candidate $C^*$ in $M_\beta$, or $\varnothing$

  \vspace{3pt}
  \State \textbf{// Step 1: Retrieve source signature}
  \State Retrieve $\tau_{\mathrm{arch},\alpha}$ and $\mathcal{S}_\alpha$ from $\mathcal{R}$

  \vspace{3pt}
  \State \textbf{// Step 2: Identify structural candidates}
  \State $\delta_\rho \gets 1/L_\beta$
         \Comment{one-layer depth tolerance, scales with model}
  \State $\mathcal{C}_{\mathrm{cand}} \gets \varnothing$
  \For{each stored circuit $C_\beta \in \mathcal{R}[M_\beta]$}
      \If{$\text{arch\_type}(\tau_{\mathrm{arch},\beta}) = \text{arch\_type}(\tau_{\mathrm{arch},\alpha})$
          \textbf{ and}
      \Statex \hspace{3.8em} $|\text{depth\_mean}(C_\beta)-\text{depth\_mean}(C_\alpha)|\le\delta_\rho$
          \textbf{ and}
      \Statex \hspace{3.8em} $|\text{rel\_size}(C_\beta)-\text{rel\_size}(C_\alpha)|\le\delta_{\text{size}}$}
          \State $\mathcal{C}_{\mathrm{cand}} \gets \mathcal{C}_{\mathrm{cand}} \cup \{C_\beta\}$
      \EndIf
  \EndFor

  \vspace{3pt}
  \State \textbf{// Step 3: Behavioural and causal screening}
  \State $\mathcal{C}_{\mathrm{pass}} \gets \varnothing$
  \For{each candidate $C_\beta \in \mathcal{C}_{\mathrm{cand}}$}
      \If{$\mathrm{Acc}_T(C_\beta) \ge \theta_{\mathrm{behav}}$
          \textbf{ and } $\Delta_T(C_\beta) \ge \theta_{\mathrm{causal}}$}
          \State $\mathcal{C}_{\mathrm{pass}} \gets \mathcal{C}_{\mathrm{pass}} \cup \{C_\beta\}$
      \EndIf
  \EndFor

  \vspace{3pt}
  \State \textbf{// Step 4: Select best candidate}
  \If{$\mathcal{C}_{\mathrm{pass}} \neq \varnothing$}
      \State $C^* \gets \arg\max_{C_\beta \in \mathcal{C}_{\mathrm{pass}}} \Delta_T(C_\beta)$
      \Return $C^*$
  \EndIf

  \vspace{3pt}
  \State \textbf{// Fallback: full rediscovery with compatibility check}
  \State $C' \gets$ circuit discovery on $M_\beta$ using $\mathcal{D}_T^\beta$
  \State Compute $\mathcal{L}(C')$ via Algorithm~\ref{alg:characterisation}
  \If{$d\!\left(\mathcal{S}_{\mathrm{ILP}}(\tau'),\mathcal{S}_{\mathrm{ILP}}(\tau_{\mathrm{arch},\alpha})\right)
      < \delta_{\mathrm{ILP}}$}
      \Return $C'$
      \Comment{compatible mechanism found via rediscovery}
  \Else
      \Return $\varnothing$
      \Comment{mechanisms are structurally incompatible}
  \EndIf
  \end{algorithmic}
  \end{algorithm}

\end{document}